\documentclass[twocolumn]{svjour3}     
\journalname{}
\date{}
\usepackage{graphicx}
\usepackage{amsmath}
\usepackage{amssymb}

\usepackage{math}
\usepackage{balance}
\title{Object Pose: The Link between Weak Perspective, Paraperspective, and Full Perspective\thanks{This work has been supported by the Esprit programme through the Basic Research Project SECOND (Esprit-BRA No. 6769). R. Horaud acknowledges partial support from the University of Bern, Institut fur Informatik und angewandte Mathematik, where he was a visiting scientist during July 1994.}}
\titlerunning{Object Pose: The Link between Weak Perspective, Paraperspective, and Full Perspective}
\author{Radu Horaud \and Fadi Dornaika \and Bart Lamiroy \and St\'ephane Christy}
\authorrunning{R. Horaud, F. Dornaika, B. Lamiroy \& S. Christy}
\institute{
INRIA Rh\^one-Alpes \& GRAVIR-IMAG\\
655, avenue de l'Europe\\
38330 Montbonnot Saint-Martin, FRANCE
}

\begin{document}
\maketitle

\begin{abstract}
Recently, DeMenthon \& Davis \cite{DementhonDavis92}, \cite{DementhonDavis95} proposed a method for
determining the pose of a 3-D object with respect to a camera from 3-D to
2-D point correspondences. The method consists of iteratively improving the
pose computed with a weak perspective camera model to converge, at the
limit, to a pose estimation computed with a perspective camera model.
In this paper we give an algebraic derivation of DeMenthon and Davis' method
and we show that it belongs to a larger class of methods where the
perspective camera model is approximated either at zero order (weak
perspective) or first order (paraperspective). 
We describe in detail an iterative paraperspective pose computation method for
both non coplanar and coplanar object points.
We analyse the convergence
of these methods and we conclude that the iterative paraperspective method
(proposed in this paper)
has better convergence properties than the iterative weak perspective
method. We introduce a simple way of taking into account the orthogonality
constraint associated with the rotation matrix. We analyse the sensitivity
to camera calibration errors and we define the optimal experimental setup
with respect to imprecise camera calibration. We compare the results
obtained with this method and with a non-linear optimization method.
\end{abstract}
\section{Introduction}
\label{section:Introduction}
Recently, DeMenthon \& Davis \cite{DementhonDavis92}, \cite{DementhonDavis95} 
proposed a method for
determining the pose of a 3-D object with respect to a camera from 3-D to
2-D point correspondences. The method consists of iteratively improving the
pose computed with a weak perspective camera model to converge, at the
limit, to a pose estimation computed with a perspective camera model.
To our knowledge, the method proposed by DeMenthon \& Davis is among one of the first attempts
to link linear techniques, associated with the weak perspective camera model,
with non-linear techniques, associated with a perspective camera model. Indeed, on one
side, linear resolution methods can be used but the solution thus obtained is
just an approximation and, on the other side, non linear resolution methods
lead to a very accurate solution but proper initialization is required.

One possibility could be to use a robust numerical method to perform the non
linear minimization that is initialized with the solution obtained by a linear
method. However, there are several drawbacks. First, such an approach does not take into account the simple
mathematical link that exists between the perspective model and its linear
approximations. Second, the linear approximation may be quite faraway from the
true solution and a large number of iterations would be required before the
non-linear minimization process converges to a stable solution. For example,
as it will be described below, a non-linear method requires 10 times more
iterations than the method of DeMenthon \& Davis and than the method described in
this paper.

The perspective projection is modelled by a projective transformation mapping the
3-D projective space to the 2-D projective plane. Weak perspective is just an 
affine approximation of full perspective. More precisely, it may well be viewed as a zero-order approximation: $1/(1+\varepsilon ) \approx 1$. Paraperspective 
\cite{Aloimonos90}
is a first order approximation of full perspective: $1/(1+\varepsilon ) 
\approx 1-\varepsilon$. The method proposed by DeMenthon \& Davis starts with computing the pose of an object using weak perspective and after a few iterations
converges towards a pose estimated under perspective. The method is very 
elegant, very fast, and quite accurate. It is however limited to situations 
where the weak perspective approximation is valid. If the object is close to the camera and/or at some distance away from the optical axis then the pose algorithm of DeMenthon \& Davis either converges very slowly (100 iterations rather
than 5 to 10) or it doesn't converge at all.

In this paper we show how the initial method proposed in \cite{DementhonDavis92} and \cite{DementhonDavis95} may be extended to paraperspective. More precisely, we describe a method for computing object pose with a paraperspective model and we establish the link between paraperspective pose and perspective pose.
We show that the case of a coplanar set of object points is somehow more
complex than the case of a non coplanar set of object points and we describe a
method which can deal with both cases.
We show both theoretically and experimentally that our method has better convergence properties than the method proposed by DeMenthon \& Davis. Moreover we introduce a simple computational way of taking into account the orthogonality constraint associated with the 3$\times$3 matrix describing the 
orientation between the 3-D object and the camera. 
Indeed, the linear pose algorithms using weak perspective and paraperspective do not guarantee that this matrix is orthogonal. The orthogonalization method that we describe below computes the best rotation in close form using unit quaternions. This orthogonalization method considerably increases the accuracy of the method at the cost of very few extra computations. We characterize the best experimental setup that allows one to compute a precise pose even in the presence of camera calibration errors. Finally we provide a comparison between the results obtained with this method and the results obtained with a non-linear minimization method
\cite{PhongHoraud93b}, \cite{PhongHoraud95}.

\subsection{Background}
The problem of object pose from 2-D to 3-D correspondences has received a lot 
of attention both in the photogrammetry and computer vision literatures. 
Haralick et al. \cite{Haralicketal89} cites a German dissertation from 1958
that surveys nearly 80 different solutions from the photogrammetry literature.
Various approaches to the object pose (or external camera parameters) problem
fall into 2 distinct categories: closed-form solutions and numerical
solutions. 

Closed-form solutions may be applied only to a limited number of
correspondences \cite{FischlerBolles81}, 
\cite{HoraudConioLeboulleuxLacolle89}, \cite{Dhome89}. Whenever the number of
correspondences is larger than 4 then 
closed-form solutions are not efficient any
more and iterative numerical solutions are necessary \cite{Yuan89},
\cite{Lowe91}, \cite{Haralicketal89}. These approaches are, in general, very
robust but they converge towards the correct solution on the premise that a good
initial estimate of the true solution is provided. Phong et al.
\cite{PhongHoraud95} describe a method that uses trust-region optimisation and
that is less sensitive to initialisation than other minimization methods.
However, the method of Phong et al. performs well for a relatively large
number of correspondences. Whenever the number of correspondences is between 3
and 10, then the trust-region minimization method either requires a large
number of iterations or doesn't converge towards the correct solution. From a
practical point of view, it is important to have available an object pose algorithm
which doesn't necessarily require a large number of correspondences and which
doesn't suffer from the limitations that are inherent to closed-form methods.

The method recently proposed by DeMenthon \& Davis 
\cite{DementhonDavis92}, \cite{DementhonDavis95} requires an arbitrary
number of point correspondences. The 3-D points must be in 
general position
(i.e., non coplanar)
and there should be at least 4 point correspondences. The
method is fast and it is robust with respect to image measurements and to
camera calibration errors. However, the object pose algorithm as suggested by
DeMenthon and Davis has convergence problems if the object is too faraway from
the optical axis of the camera --- a limitation that can be overcome as
explained farther in this paper. The method of DeMenthon \& Davis was extended
to the case of a coplanar set of points by Oberkampf, DeMenthon, and Davis
\cite{OberkampfDementhonDavis93} and by DeMenthon \cite{daniel}.


\subsection{Paper organization}
The remainder of this paper is organized as follows. In Section \ref{section:Camera-models} we briefly recall the perspective camera model and two possible
approximations of this model, weak and paraperspective.
Section~\ref{section:From-weak-perspective} describes the iterative weak
perspective algorithm as it has been proposed in \cite{DementhonDavis95}.
Section~\ref{section:From-paraperspective} describes the new iterative paraperspective 
algorithm that we propose as well as how to compute the object
pose from a paraperspective approximation. 
Section~\ref{section:Solving-the-linear-equations} provides details for
solving the linear equations associated with pose computation in the case
of both non coplanar and planar sets of object points.
Section~\ref{section:An-analysis-of-convergence} analyses the convergence of both the 
iterative weak and paraperspective
algorithms and Section~\ref{section:The-orthogonality-constraint}
shows how to improve the pose algorithms using a simple and straightforward
formulation of the orthogonality constraint. Section~\ref{section:Sensitivity-to-camera-calibration} analyses the sensitivity of both algorithms with respect to camera
calibration errors, and Section~\ref{section:Experiments} provides an experimental
comparison of the two methods described in this paper together with a
comparison of these methods with a non-linear minimization method. 
Section~\ref{section:Examples} shows many examples of application of pose
computation.
Finally, Section~\ref{section:Discussion} provides a discussion and gives directions for
future work.
\section{Camera models}
\label{section:Camera-models}
\begin{figure}
\centering
\includegraphics[width=0.95\columnwidth]{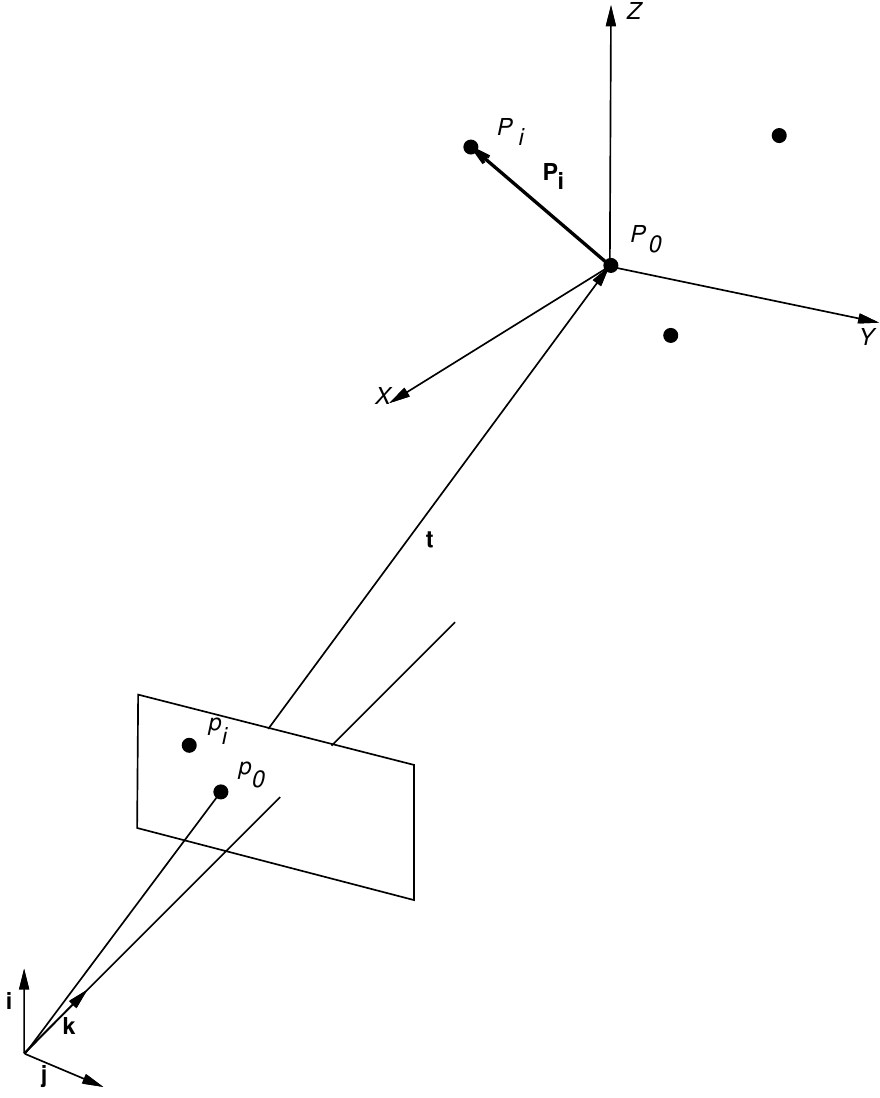}
\caption{This figure shows the general setup. One of the object's point is
selected as the origin of the object frame. Therefore, the pose parameters are
the coordinates of this object point ($P_0$) in the camera frame and the
orientation of the object frame with respect to the camera frame.}
\label{fig:setup}
\end{figure}

We consider a simple setup, as depicted on Figure \ref{fig:setup}. We denote by $P_i$ a 3-D point with coordinates $X_i$, $Y_i$, and $Z_i$ in a frame 
that is attached to the object -- the object frame. The origin of this frame is
the object point $P_0$. An object point $P_i$ projects onto the image in
$p_i$ with camera
coordinates $x_i$ and $y_i$ and we have ($\Pvect _i$ is the vector
from point  $P_0$ to point  $P_i$):
\begin{eqnarray}
\label{eq:x-perspective}
x_i & = & \frac{\ivect \cdot \Pvect _i + t_x}{\kvect \cdot \Pvect _i + t_z} \\
\label{eq:y-perspective}
y_i & = & \frac{\jvect \cdot \Pvect _i + t_y}{\kvect \cdot \Pvect _i + t_z}
\end{eqnarray}
These equations describe the classical perspective camera model where the
rigid transformation from the object frame to the camera frame is:
\[
T = \left( \begin{array}{cc}
	\ivect ^T & t_x \\
	\jvect ^T & t_y \\
	\kvect ^T & t_z \\
	0\; 0 \; 0 & 1
    \end{array} \right)
=
\left( \begin{array}{cc}
R & \tvect \\
0\; 0 \; 0 & 1
    \end{array} \right)
\]
The relationship between the camera coordinates and the image coordinates may
be easily obtained by introducing the intrinsic camera parameters:
\begin{eqnarray}
\label{eq:u-intrinsic}
u_i & = & \alpha _u x_i + u_c \\
\label{eq:v-intrinsic}
v_i & = & \alpha _v y_i + v_c
\end{eqnarray}

In these equations $\alpha_u$ and $\alpha_v$ are the vertical and horizontal
scale factors and $u_c$ and $v_c$ are the image coordinates of the
intersection of the optical axis with the image plane.

 We divide both the numerator and the denominator of eqs.
(\ref{eq:x-perspective}) and (\ref{eq:y-perspective}) by $t_z$. We introduce
the following notations:
\begin{itemize}
\item $\Ivect = \ivect / t_z$ is the first row of the rotation matrix scaled
by the z-component of the translation vector;
\item $\Jvect = \jvect / t_z$ is the second row of the rotation matrix scaled
by the z-component of the translation vector;
\item $x_0 = t_x / t_z$ and $y_0 = t_y / t_z$ are the camera coordinates of
$p_0$ which is the projection of $P_0$ -- the origin of the object frame, and
\item $\varepsilon _i = \kvect \cdot \Pvect _i / t_z$.
\end{itemize}

We may now rewrite the perspective equations as:
\begin{eqnarray}
\label{eq:x-eps-perspective}
x_i & = & \frac{\Ivect \cdot \Pvect _i + x_0}{1+\varepsilon _i}\\
\label{eq:y-eps-perspective}
y_i & = & \frac{\Jvect \cdot \Pvect _i + y_0}{1+\varepsilon _i}
\end{eqnarray}
These equations may also be written as in \cite{DementhonDavis95}:
\begin{eqnarray}
\label{eq:x-dem-perspective}
x_i (1+\varepsilon _i ) - x_0 & =&  \Ivect \cdot \Pvect _i \\
\label{eq:y-dem-perspective}
y_i (1+\varepsilon _i ) - y_0 & =&  \Jvect \cdot \Pvect _i 
\end{eqnarray}

Whenever the object is at some distance from the camera, the $\varepsilon _i$ are
small compared to 1. We may therefore introduce two approximations of the
perspective equations: weak and paraperspective. One can find a geometric description of these projections in \cite{poelman}.

\subsection{Weak perspective}
\begin{figure}[thb]
\centering
\includegraphics[width=0.95\columnwidth]{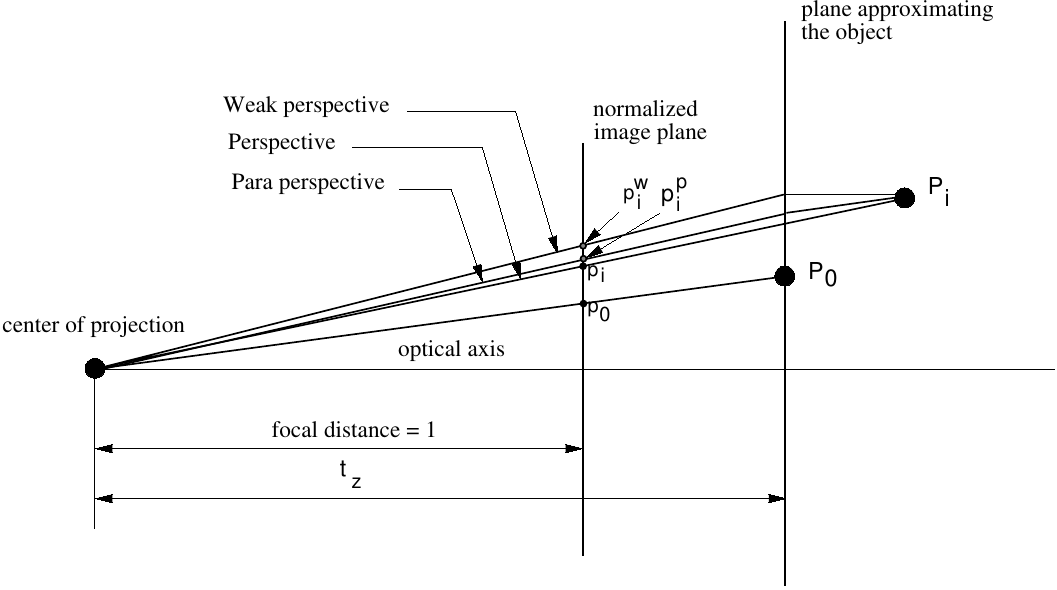}
\caption{This figure shows $p_0$, $p_i$ -- the perspective projections of two
object points,
$P_0$ and $P_i$, and $p_i^w$, $p_i^p$ -- the weak and paraperspective
projections of $P_i$. The
quality of the paraperspective projection depends on the angle between
the line of projection of $P_i$ and the line of projection of $P_0$.
Similarly, the quality of the weak perspective projection depends on the angle
between the line of projection of $P_i$ and the optical axis. }
\label{fig:weakpara-projections}
\end{figure}

Weak perspective assumes that the object points lie in a plane parallel
to the image plane passing through the origin of the object frame, i.e.,
$P_0$, e.g.,
Figure~\ref{fig:weakpara-projections}. 
This is equivalent to a zero-order approximation:
\[	\frac{1}{1+\varepsilon _i} \approx 1 \; \; \forall i, \; i \in \{1...n\}
\]
With this approximation, eqs. (\ref{eq:x-eps-perspective}) and
(\ref{eq:y-eps-perspective}) become:
\begin{eqnarray}
\label{eq:x-dem-weak}
x_i^w - x_0 & =&  \Ivect \cdot \Pvect _i \\
\label{eq:y-dem-weak}
y_i^w - y_0 & =&  \Jvect \cdot \Pvect _i 
\end{eqnarray}
In these two equations $x_i^w$ and $y_i^w$ are the camera coordinates of the weak perspective
projection of the point $P_i$. By identification with eqs.
(\ref{eq:x-dem-perspective}) and
(\ref{eq:y-dem-perspective}) we obtain the relationship between the weak
perspective and the perspective projections of $P_i$:
\begin{eqnarray}
\label{eq:xi-weak-persp}
x_i^w & =& x_i (1+\varepsilon _i )\\
\label{eq:yi-weak-persp}
y_i^w & =& y_i (1+\varepsilon _i )
\end{eqnarray}

These equations allow us to determine the quality of the weak perspective
approximation with respect to the perspective projection. Indeed the error
between the weak perspective projection and the ``true" projection is:
\begin{eqnarray}
\label{eq:xi-weak-error}
\Delta x^w & = & | x_i^w - x_i | = | x_i \varepsilon_i | \\
\label{eq:yi-weak-error}
\Delta y^w & = & | y_i^w - y_i | = | y_i \varepsilon_i | 
\end{eqnarray}
Hence the quality of the approximation depends both on the value of
$\varepsilon_i$ {\sc and} on the position of the point in the image. We
consider for example a 512$\times$512 image. Approximate values for
the intrinsic parameters are: $\alpha_u=\alpha_v=1000$ and 
$u_c=v_c=256$. Therefore using eq.~(\ref{eq:u-intrinsic}) and
(\ref{eq:v-intrinsic}) we have:
\[	0 \leq x_i \; , \; y_i \leq 0.25	\]

The lower bound corresponds to the point of intersection of the optical axis
with the image plane ($x_i = y_i = 0$). The upper bound corresponds to the
image borders: 
\[	x_i = \frac{u_i - u_c}{\alpha _u} = \frac{512 - 256}{1000} \approx
0.25 \]

We conclude that for small distance/size ratios, i.e., large values for $\varepsilon_i$, 
the weak
perspective approximation is still valid {\em
provided that the object lies in the
neighbourhood of the optical axis}.

\subsection{Paraperspective}
Paraperspective may almost be viewed as a first-order approximation of
perspective. Indeed, two approximations are needed in order to obtain
the paraperspective camera model. With the approximation:
\[ \frac{1}{1+\varepsilon _i} \approx 1-\varepsilon _i \; \; \forall i, \; i \in \{1...n\}
\]
we obtain the paraperspective projection of $P_i$, e.g.,
Figure~\ref{fig:weakpara-projections}:
\begin{eqnarray*}
x_i^p & = & (\Ivect \cdot \Pvect _i + x_0)( 1-\varepsilon _i)\\
    & \approx & \Ivect \cdot \Pvect _i + x_0 - x_0 \varepsilon _i\\
    & = & \frac{\ivect \cdot \Pvect _i}{t_z} + x_0 - x_0 \frac{\kvect \cdot
\Pvect _i}{t_z}
\end{eqnarray*}
where the term in $1/t_z^2$ was neglected. 
There is a similar expression for $y_i^p$.

Finally, the paraperspective equations are:
\begin{eqnarray}
\label{eq:x-dem-para}
x_i^p - x_0 & = &  \frac{\ivect - x_0\; \kvect}{t_z} \cdot \Pvect _i \\
\label{eq:y-dem-para}
y_i^p - y_0 & = &  \frac{\jvect - y_0\; \kvect}{t_z} \cdot \Pvect _i 
\end{eqnarray}
In order to obtain the relationship between the paraperspective and the
perspective projections of $P_i$ we can write these equations as follows:
\begin{eqnarray*}
x_i^p & = & x_0 + \Ivect \cdot \Pvect_i - x_0 \varepsilon _i\\
y_i^p & = & y_0 + \Jvect \cdot \Pvect_i - y_0 \varepsilon _i\\
\end{eqnarray*}
Again, by identification with eqs.
(\ref{eq:x-dem-perspective}) and
(\ref{eq:y-dem-perspective}) we obtain the relationship between the 
paraperspective and the perspective projections of $P_i$:
\begin{eqnarray}
\label{eq:xi-para-persp}
x_i^p & = & x_i (1+\varepsilon _i ) - x_0 \varepsilon _i\\
\label{eq:yi-para-persp}
y_i^p & = & y_i (1+\varepsilon _i ) - y_0 \varepsilon _i
\end{eqnarray}

As in the previous section, we can easily estimate the error between the
paraperspective and perspective projections:
\begin{eqnarray}
\label{eq:xi-para-error}
\Delta x^p & = & | x_i^p - x_i | = | (x_i - x_0) \varepsilon_i | \\
\label{eq:yi-para-error}
\Delta y^p & = & | y_i^p - y_i | = | (y_i - y_0) \varepsilon_i | 
\end{eqnarray}

Whenever an object point $P_i$ is far from the optical axis then the weak perspective
model is a poor approximation. However, a proper choice of the origin, i.e,
$P_0$, and the use of the paraperspective model can compensate and provide a
good approximation even if $\varepsilon_i$ is not small.
\section{From weak perspective to perspective}
\label{section:From-weak-perspective}

In order to solve the pose problem, DeMenthon \& Davis \cite{DementhonDavis95}
noticed that eqs. (\ref{eq:x-dem-weak}) and (\ref{eq:y-dem-weak}) are similar
to eqs. (\ref{eq:x-dem-perspective}) and (\ref{eq:y-dem-perspective}) for
which the $\varepsilon _i$ are set to 0. They conclude that:
\begin{itemize}
\item Whenever the $\varepsilon _i$ are fixed (not necessarily null) the pose
equations (\ref{eq:x-dem-perspective}) and (\ref{eq:y-dem-perspective})
become linear in $\Ivect$ and $\Jvect$. A solution can be found if 
at least 4 object points are provided 
(see Section~\ref{section:Solving-the-linear-equations}).
\item It is possible to solve eqs. (\ref{eq:x-dem-perspective}) and
(\ref{eq:y-dem-perspective}) {\em iteratively} by successive linear
approximations. 
\end{itemize}
In this case the pose algorithm proposed in \cite{DementhonDavis95}
starts with a weak perspective camera model
and computes an approximated pose. This approximated pose is improved
iteratively as follows:
\begin{enumerate}
\item For all $i$, $i \in \{1...n\}$, $n \geq 3$,  $\varepsilon _i = 0$;
\item Solve the overconstrained linear system of equations (\ref{eq:x-dem-perspective}) and (\ref{eq:y-dem-perspective}) which provides an estimation of
vectors $\Ivect$ and $\Jvect$, i.e.,
Section~\ref{section:Solving-the-linear-equations};
\item Compute the position and orientation of the object frame with respect to
the camera frame:
\begin{eqnarray*}
t_z & = & \frac{1}{2}\left( \frac{1}{\| \Ivect \|} + \frac{1}{\| \Jvect
\|}\right)\\
t_x & = & x_0 \; t_z \\
t_y & = & y_0 \; t_z \\
\ivect & = & \frac{\Ivect}{\| \Ivect \|}\\
\jvect & = & \frac{\Jvect}{\| \Jvect \|}\\
\kvect & = & \ivect \times \jvect
\end{eqnarray*}
\item For all $i$, compute:
\[	\varepsilon _i = \frac{\kvect \cdot \Pvect _i}{t_z}	\]
If the $ \varepsilon _i$ computed at this iteration are equal to the
 $ \varepsilon _i$ computed at the previous iteration then stop the procedure, 
otherwise go to step 2.
\end{enumerate}

While iterating, the above algorithm replaces the actual image coordinates
$x_i$ and $y_i$ (which are perspective projections of a 3-D point) 
with $x_i(1+\varepsilon_i)$ and $y_i(1+\varepsilon_i)$, which
are weak perspective projections of the same 3-D point. 
A geometric interpretation of this algorithm 
can be found in \cite{daniel}, \cite{DementhonDavis95}.

\section{From paraperspective to perspective}
\label{section:From-paraperspective}

In this section we provide a generalization of the above algorithm to deal
with the paraperspective case. We consider again the perspective equations
(\ref{eq:x-dem-perspective}) and
(\ref{eq:y-dem-perspective}) and let us subtract the {\em paraperspective
term} from both the left and right sides of these equations. We obtain:
\begin{eqnarray*}
x_i (1+\varepsilon _i ) - x_0 - x_0 \underbrace{\frac{1}{t_z} \kvect \cdot \Pvect
_i}_{\varepsilon _i} & = &  
\frac{1}{t_z} \ivect \cdot \Pvect _i  - x_0 \frac{1}{t_z} \kvect \cdot \Pvect
_i \\
y_i (1+\varepsilon _i ) - y_0 - y_0 \underbrace{\frac{1}{t_z} \kvect \cdot
\Pvect _i}_{\varepsilon _i} & = &  
\frac{1}{t_z} \jvect \cdot \Pvect _i  - y_0 \frac{1}{t_z} \kvect \cdot \Pvect
_i
\end{eqnarray*}

These equations can be written more compactly as:
\begin{eqnarray}
\label{eq:x-radu-para}
(x_i - x_0)(1+\varepsilon _i ) & = & \Ivect _p \cdot \Pvect _i \\
\label{eq:y-radu-para}
(y_i - y_0)(1+\varepsilon _i ) & = & \Jvect _p \cdot \Pvect _i 
\end{eqnarray}
with:
\begin{eqnarray}
\label{eq:Ip}
\Ivect _p & = & \frac{\ivect - x_0 \; \kvect}{t_z} \\
\label{eq:Jp}
\Jvect _p & = & \frac{\jvect - y_0 \; \kvect}{t_z} 
\end{eqnarray}

One may notice that when all the $\varepsilon _i$ are null, the
perspective equations above -- eqs. (\ref{eq:x-radu-para}) and
(\ref{eq:y-radu-para}) -- become identical to the paraperspective equations
-- eqs. (\ref{eq:x-dem-para}) and (\ref{eq:y-dem-para}). 

As in the weak perspective case, we have now a linear method for computing the
pose with a paraperspective model and an iterative method to compute the
object pose with a perspective model by successive paraperspective
approximations. More precisely, the pose parameters can be derived from 
$\Ivect _p$ and $\Jvect _p$ as follows.

First, one may notice that:
\begin{eqnarray*}
\| \Ivect _p \| ^2 & = & \frac{(\ivect - x_0 \; \kvect) \cdot (\ivect - x_0 \;
\kvect)}{t_z^2}\\
		   & = & \frac{1+x_0^2}{t_z^2}
\end{eqnarray*}
and:
\[ \| \Jvect _p \| ^2 = \frac{1+y_0^2}{t_z^2} \]
We therefore obtain:
\begin{equation}
t_z = \frac{1}{2}\left(\frac{\sqrt{1+x_0^2}}{\|\Ivect _p\|} +
\frac{\sqrt{1+y_0^2}}{\|\Jvect _p\|} \right) 
\label{eq:tz-para}
\end{equation}
and:
\[ t_x = x_0 t_z \]
\[ t_y = y_0 t_z \]

Second, we derive the three orthogonal unit vectors $\ivect$, $\jvect$, and
$\kvect$. From eqs. (\ref{eq:Ip}) and (\ref{eq:Jp}) we obtain:
\begin{eqnarray}
\label{eq:i-para}
\ivect & = & t_z \; \Ivect _p + x_0 \; \kvect \\
\label{eq:j-para}
\jvect & = & t_z \; \Jvect _p + y_0 \; \kvect 
\end{eqnarray}
The third vector, $\kvect$ is the cross-product of these two vectors:
\begin{eqnarray*}
\kvect & = & \ivect \times \jvect \\
       & = &  t_z ^2 \:\Ivect _p \times \Jvect _p + t_z y_0\: \Ivect _p \times
\kvect - t_z x_0 \:\Jvect _p \times \kvect
\end{eqnarray*}
Let $S(\avect)$ be the skew-symmetric matrix associated with a 3-vector
$\avect$ and $I_{3\times 3}$ be the identity matrix. 
The previous expression can now be written as follows:
\begin{equation}
\left( I_{3\times 3} - t_z y_0 \:S(\Ivect _p) + t_z x_0 \:S(\Jvect _p)
\right) \; \kvect = t_z ^2 \: \Ivect _p \times \Jvect _p
\label{eq:k-para}
\end{equation}

This equation allows us to compute $\kvect$, provided that the linear system
above has full rank. Indeed, the 3$\times$3 matrix $A$:
\[ A = I_{3\times 3} - t_z y_0 \:S(\Ivect _p) + t_z x_0 \:S(\Jvect _p) \]
is of the form:
\[ A = \left( \begin{array}{rrr}
	  1 & c & -b \\
	 -c & 1 &  a \\
	  b & -a & 1
	  \end{array}
   \right) 
\]
Its determinant is always strictly positive:
\[ det(A) = 1 + a^2 + b^2 + c^2 \]
Therefore, one can easily determine $\kvect$ using eq.~(\ref{eq:k-para}) and 
$\ivect$ and $\jvect$ using eqs.~(\ref{eq:i-para}) and (\ref{eq:j-para}).

The $\varepsilon _i$ can now be easily computed as before and the pose algorithm
becomes:
\begin{enumerate}
\item For all $i$, $i \in \{1...n\}$, $n \geq 3$,  $\varepsilon _i = 0$;
\item Solve the overconstrained linear system of equations
(\ref{eq:x-radu-para}) and (\ref{eq:y-radu-para}) which provides an estimation of vectors $\Ivect _p$ and $\Jvect _p$, i.e.,
Section~\ref{section:Solving-the-linear-equations};
\item Compute the position ($t_x$, $t_y$, and $t_z$) and orientation 
($\ivect$, $\jvect$, and $\kvect$) of the object frame with respect to
the camera frame as explained above in this Section;
\item For all $i$, compute:
\[      \varepsilon _i = \frac{\kvect \cdot \Pvect _i}{t_z}      \]
If the $ \varepsilon _i$ computed at this iteration are equal to the
 $ \varepsilon _i$ computed at the previous iteration then stop the procedure,
otherwise go to step 2.
\end{enumerate}

\section{Solving the linear equations}
\label{section:Solving-the-linear-equations}

Both the weak perspective and paraperspective iterative algorithms need to
solve an overconstrained linear system of equations, namely eqs.
(\ref{eq:x-dem-perspective}), (\ref{eq:y-dem-perspective})  (weak perspective)
and eqs.
(\ref{eq:x-radu-para}), (\ref{eq:y-radu-para}) (paraperspective). In matrix
form these equations can be written as:
\begin{eqnarray}
\label{eq:linear-x}
\underbrace{P}_{n\times 3} \underbrace{\Ivect}_{3\times 1} & = & 
\underbrace{\xvect}_{n\times 1} \\
\label{eq:linear-y}
\underbrace{P}_{n\times 3} \underbrace{\Jvect}_{3\times 1} & = & 
\underbrace{\yvect}_{n\times 1} 
\end{eqnarray}
where $P$ is a ${n\times 3}$ matrix formed by the 3-D coordinates of $n$
vectors $\Pvect_1\:\ldots\: \Pvect_n$. Since the point $P_0$ is the origin of the object frame, this matrix can be written as~:
\[
P = \left( \begin{array}{ccc}

X_1    & Y_1     & Z_1    \\
\vdots & \vdots & \vdots \\
X_n   &  Y_n   & Z_n  
\end{array} \right)
\]

In order to solve for these linear equations one has to distinguish two cases:
non coplanar and coplanar sets of object points.

\subsection{Non coplanar object points}

If the object points are not coplanar, the rank of $P$ is 3 and therefore the
solutions for $\Ivect$ and $\Jvect$ (and equivalently for $\Ivect_p$ and 
$\Jvect_p$) are simply given by:
\begin{eqnarray*}
\Ivect & = & (P^TP)^{-1}P^T \: \xvect \\
\Jvect & = & (P^TP)^{-1}P^T \: \yvect 
\end{eqnarray*}
One may notice that the pseudo-inverse of $P$ can be computed off-line and
hence the estimation of $\Ivect$ and $\Jvect$ is particularly efficient.

\subsection{Coplanar object points}

\begin{figure}[tbh]
\centering
\includegraphics[width=0.95\columnwidth]{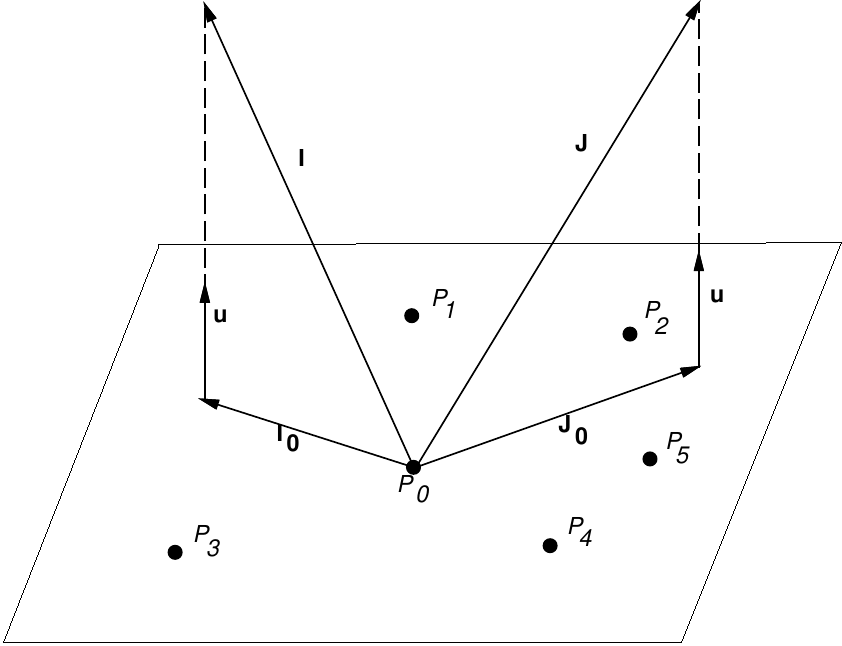}
\caption{A coplanar configuration of object points.}
\label{fig:coplanar-points}
\end{figure}

If the object points are coplanar then the rank of $P$ is 2 and the above
solution cannot be considered anymore. We consider the plane formed in this
case by the object points and let $\uvect$ be the unit vector orthogonal to
this plane. Vectors $\Ivect$ and $\Jvect$ (and equivalently $\Ivect_p$ and
$\Jvect_p$) can be written as a sum of a vector belonging to this plane and a
vector perpendicular to this plane, namely (see \cite{daniel},
\cite{OberkampfDementhonDavis93} and
Figure~\ref{fig:coplanar-points}):
\begin{eqnarray}
\label{eq:I-decomposition}
\Ivect & = & \Ivect_0 + \lambda \: \uvect \\
\label{eq:J-decomposition}
\Jvect & = & \Jvect_0 + \mu\: \uvect 
\end{eqnarray}

By substituting these expressions for $\Ivect$ and $\Jvect$ into eqs. 
(\ref{eq:linear-x}) and (\ref{eq:linear-y}) we obtain:
\begin{eqnarray*}
P \:\Ivect_0 & = & \xvect \\
P \:\Jvect_0 & = & \yvect
\end{eqnarray*}
These linear equations can be solved provided that the following additional
linear constraints are used:
\[
\uvect \cdot \Ivect_0 = 0
\]
\[
\uvect \cdot \Jvect_0 = 0
\]
Thus we obtain solutions for $\Ivect_0$ and $\Jvect_0$ as follows.
\begin{eqnarray*}
\Ivect_0 & = & (P'^TP')^{-1}P'^T \: \left( \begin{array}{c} \xvect \\ 0
\end{array} \right) \\
\Jvect_0 & = & (P'^TP')^{-1}P'^T \: \left( \begin{array}{c} \yvect \\ 0
\end{array} \right) 
\end{eqnarray*}
With $P'$ defined by:
\[
P' = \left( \begin{array}{c} P \\ \uvect \end{array} \right)
\]
Obviously, the rank of $P'$ is equal to 3.

In order to estimate $\Ivect$ and $\Jvect$ (and equivalently $\Ivect_p$ and
$\Jvect_p$) one is left with the estimation of two scalars, $\lambda$ and
$\mu$. In the case of weak perspective Oberkampf, DeMenthon, and Davis
\cite{OberkampfDementhonDavis93} provided a solution using the constraints
$\|\Ivect\|=\|\Jvect\|$ and $\Ivect\cdot\Jvect =0$.

In the case of paraperspective a similar solution can be obtained using the
following constraints onto $\Ivect_p$ and
$\Jvect_p$ (derived from eqs. (\ref{eq:Ip}) and (\ref{eq:Jp})):
\begin{eqnarray*}
\| \Ivect _p \|^2 & = & \frac{1+ x_0^2}{t_z^2} \\
\| \Jvect _p \|^2 & = & \frac{1+ y_0^2}{t_z^2} \\
\Ivect _p \cdot \Jvect _p & = & \frac{x_0 y_0}{t_z^2}
\end{eqnarray*}
By eliminating $t_z$ we obtain two constraints:
\begin{eqnarray*}
\Ivect _p \cdot \Jvect _p & = & \frac{x_0 y_0}{1+ x_0^2}\:\| \Ivect _p \|^2 \\
\Ivect _p \cdot \Jvect _p & = & \frac{x_0 y_0}{1+ y_0^2}\:\| \Jvect _p \|^2 
\end{eqnarray*}
By substituting in these expressions $\Ivect_p$ and
$\Jvect_p$ given by eqs. (\ref{eq:I-decomposition}) and 
(\ref{eq:J-decomposition}) we obtain:
\begin{eqnarray*}
\Ivect _0 \cdot \Jvect _0 + \lambda\: \mu  & = & \frac{x_0 y_0}{1+ x_0^2}
\:(\| \Ivect _0 \|^2 + \lambda^2) \\
\Ivect _0 \cdot \Jvect _0 + \lambda\: \mu  & = & \frac{x_0 y_0}{1+ y_0^2}
\:(\| \Jvect _0 \|^2 + \mu^2) 
\end{eqnarray*}
And finally, 
by eliminating $\mu$ we obtain a biquadratic equation in one unknown:
\begin{equation}
{\cal A} \:\lambda^4 + {\cal B} \:\lambda^2 + {\cal C} = 0
\label{eq:biquadratic}
\end{equation}
With:
\begin{eqnarray*}
{\cal A} & = & a^2 - g \\
{\cal B} & = & 2\,a^2d - g\,d + e - 2\,a\,c \\
{\cal C} & = & a^2d^2 + c^2 - 2\,a\,c\,d \\
a & = & \frac{x_0 y_0}{1+ x_0^2} \\
b & = & \frac{x_0 y_0}{1+ y_0^2} \\
c & = & \Ivect _0 \cdot \Jvect _0 \\
d & = & \| \Ivect _0 \|^2 \\
e & = & \| \Jvect _0 \|^2 \\
g & = & \frac{1+ y_0^2}{1+ x_0^2}
\end{eqnarray*}

In order to study the number of real roots of eq.~(\ref{eq:biquadratic}) we
substitute $\lambda^2$ by $t$:
\[
{\cal A}\:t^2 + {\cal B} \:t + {\cal C} = 0
\]
We examine the signs of the roots of this equation. Therefore we have to
examine the sign of their product, i.e., ${\cal C}/{\cal A}$. We have:
\begin{eqnarray*}
\frac{{\cal C}}{{\cal A}} & = & f(x_0,y_0,c,d) \\
 & = & \frac{-x_0^2y_0^2\;d^2 - (1+ x_0^2)^2\;c^2 + 2x_0 y_0(1+ x_0^2)\;c\,d}{1+ x_0^2+ y_0^2} \\
 & = & - \frac{(x_0y_0\;d - (1+ x_0^2)\;c)^2}{1+ x_0^2+ y_0^2}
\end{eqnarray*}

Therefore, the value of ${\cal C}/{\cal A}$ is either negative or null. 
Therefore there are one positive (or null) 
root and one negative
(or null) root for $t$. 
Hence there are two real roots for $\lambda$ -- a positive
one and a negative one -- and two imaginary roots.

\begin{figure}[htbp]
\centering
\includegraphics[width=0.95\columnwidth]{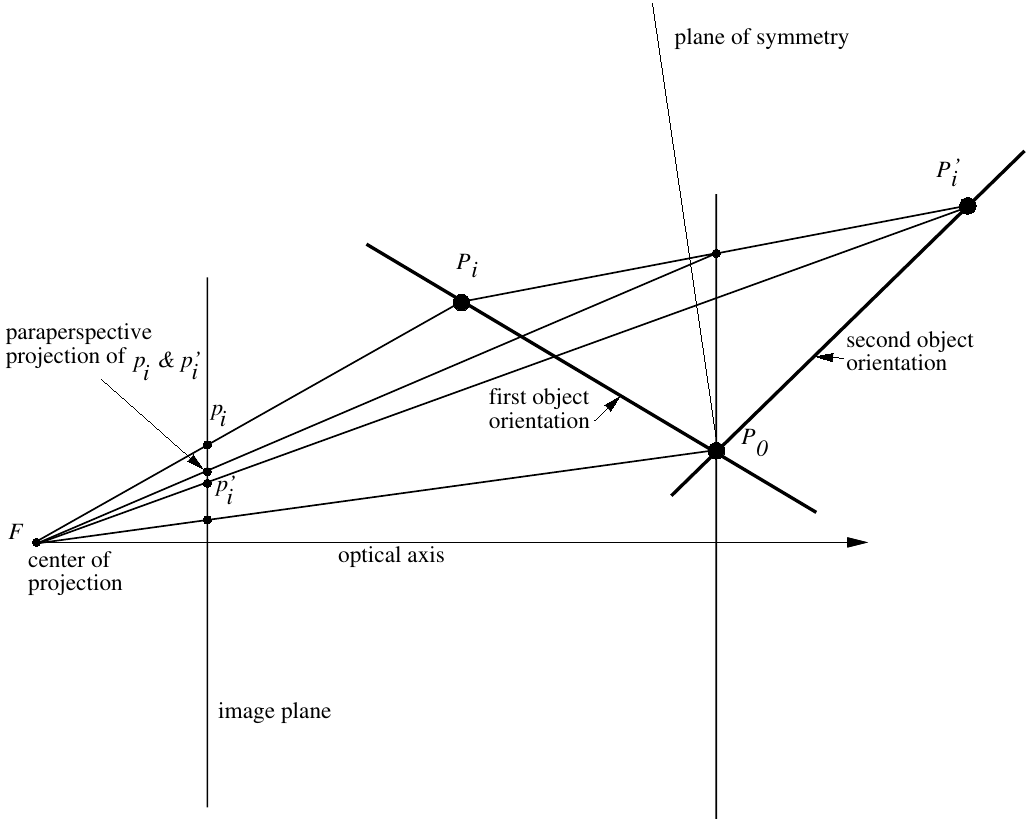}
\caption{A planar object and a paraperspective camera model produces two
orientations. However, these two object orientations have distinct perspective projections
-- $p_i$ and $p'_i$.}
\label{fig:reversal}
\end{figure}

The two real roots for $\lambda$ provide two solutions for $\mu$ 
and hence there are two solutions for $\Ivect_p$ and $\Jvect_p$. These two
solutions correspond to the well-known reversal ambiguity associated with
an affine camera model. These two solutions are shown on
Figure~\ref{fig:reversal}. The points $P_i$ (lying on the planar object in one orientation)
and $P'_i$ (lying on the planar object in another orientation) have the same paraperspective
projection but different perspective projections. Notice that these 
object orientations are symmetric with respect to a plane which is perpendicular to the
line of sight passing through $P_0$.

\begin{figure}[htbp]
\centering
\includegraphics[width=0.95\columnwidth]{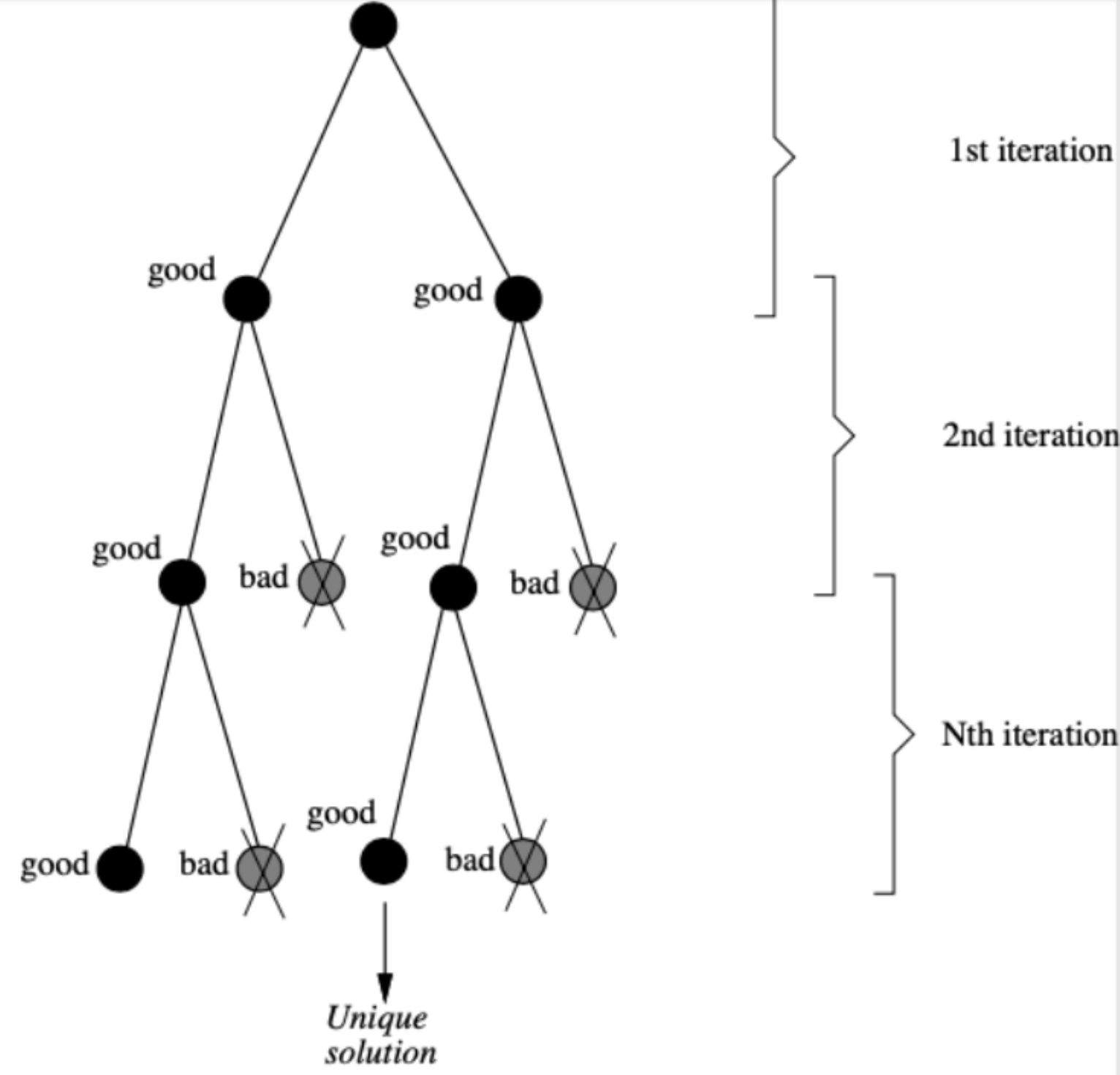}
\caption{A binary tree search for selecting the best object pose when the
object points are coplanar.}
\label{fig:searchtree}
\end{figure}

Therefore, the iterative algorithm described in Section~\ref{section:From-paraperspective} produces two poses at each iteration. The two poses have the same
translation vector but different orientations. Hence, after $n$ iterations
there will be $2^n$ solutions. In order to avoid this redundancy we proceed as
follows. At the first iteration we retain both solutions while at the next
iterations we retain only one solution -- the solution which is the most
consistent with the data. At convergence we obtain two solutions for the
orientation of the planar object: one solution associated with the ``left"
branch of the search tree and another solution associated with the ``right"
branch of the search tree (see Figure~\ref{fig:searchtree}). Among the
final two solutions, one of them is generally closer to the image data than
the other. However, if the orientation of the planar object is close to the
orientation of the plane of symmetry, the ambiguity remains and the algorithm provides
two solutions.

\section{An analysis of convergence}
\label{section:An-analysis-of-convergence}

In order to analyse the convergence of these algorithms (sections 
\ref{section:From-weak-perspective}
and \ref{section:From-paraperspective}) we consider again eqs. 
(\ref{eq:x-dem-perspective}) and (\ref{eq:y-dem-perspective}):
\begin{eqnarray*}
x_i (1+\varepsilon _i ) - x_0 & =&  \Ivect \cdot \Pvect _i \\
y_i (1+\varepsilon _i ) - y_0 & =&  \Jvect \cdot \Pvect _i
\end{eqnarray*}
as well as eqs.
(\ref{eq:x-radu-para}) and (\ref{eq:y-radu-para}):
\begin{eqnarray*}
(x_i - x_0)(1+\varepsilon _i ) & = & \Ivect _p \cdot \Pvect _i \\
(y_i - y_0)(1+\varepsilon _i ) & = & \Jvect _p \cdot \Pvect _i
\end{eqnarray*}

Both these sets of equations describe the {\em full} perspective projection.
The first ones allows to express the pose problem in terms of an iterative
weak perspective algorithm while the second ones allows to express the pose
problem in terms of an iterative paraperspective algorithm. From
Figure~\ref{fig:weakpara-projections} and Section~\ref{section:Camera-models} 
we have:
\begin{itemize}
\item $x_0$ and $y_0$ are the camera coordinates of $p_0$, the perspective projection
of $P_0$;
\item $x_i$ and $y_i$ are the camera coordinates of $p_i$, the perspective projection
of $P_i$;
\item $x_i(1+\varepsilon _i )$ and $y_i(1+\varepsilon _i )$ are the camera coordinates of $p_i^w$, the weak perspective projection of $P_i$;
\item $x_i(1+\varepsilon_i )-x_0\varepsilon_i$ and $y_i(1+\varepsilon_i)-y_0\varepsilon_i$ 
are the camera coordinates of $p_i^p$, the paraperspective projection of $P_i$;
\end{itemize}

Therefore, if good estimates for $\varepsilon _i$ ($i\in \{1...n\}$) are
available, then the perspective pose problem is reduced to solving an
overconstrained set of linear equations. These equations provide a solution
either for $\Ivect$ and $\Jvect$ or for $\Ivect_p$ and $\Jvect_p$ from which a
pose may be easily calculated. However, in practice it is rather difficult to
provide such estimates. Hence, one has to use a heuristic. The simplest
heuristic is to initialize the values of $\varepsilon _i$ to zero and to use
either the iterative weak perspective or the iterative paraperspective
algorithms.

We consider the first one of these two algorithms and let 
$\varepsilon_i^{\star}$ be the true values that the algorithm is supposed to
compute. Therefore, at the first iteration, one attempts to solve in a
least-square sense a linear set of equations of the form:
\[
\left\{
\begin{array}{ccc}
& \vdots & \\
\Ivect \cdot \Pvect _i & = & x_i - x_0 \\
\Jvect \cdot \Pvect _i & = & y_i - y_0 \\
& \vdots &
\end{array}
\right.
\]
The pose thus computed is just an approximation since one uses $x_i$ and $y_i$
instead of $x_i^w$ and $y_i^w$. Therefore the ``pose errors" are proportional
to the following ``projection errors":
\begin{equation*}
 \Delta ^w_i  =  \| p_i - p_i^w \|^2 
  =  (x_i^2 + y_i^2) | \varepsilon_i^{\star} |^2
\end{equation*}
If these projection
errors are large, the linear least-square solution provided by the
first iteration of the algorithm will be rather different than the true
solution and convergence cannot be guaranteed. DeMenthon \& Davis noticed that
the iterative weak perspective algorithm converges with values as large 
for $\varepsilon_i^{\star}$ as 1, {\em provided that the object lies in the
neighbourhood of the optical axis}. Indeed, when the object is close to the
optical axis, the camera coordinates of its projections, $x_i$ and $y_i$, are
small (the origin of the camera frame lies onto the optical axis) and they
compensate for the large values of $\varepsilon_i^{\star}$.

In theory, the iterative paraperspective algorithm is able to deal with
configurations where the weak perspective algorithm diverges. Indeed, in the
case of paraperspective, the projection errors are:
\begin{eqnarray*}
 \Delta ^p_i & = & \| p_i - p_i^p \|^2 
  =  ((x_i - x_0)^2 + (y_i - y_0)^2) | \varepsilon_i^{\star} |^2 \\
 & = & \| p_i - p_0 \|^2   | \varepsilon_i^{\star} |^2 
\end{eqnarray*}
Whenever one wants to use the paraperspective camera model, the choice for $p_0$ (and hence
for $P_0$) is crucial. The best way to choose $p_0$ is to compute the center of gravity
of all the image points and to select the image point which is the closest to this center
of gravity. Hence,
$x_i-x_0$ and
$y_i-y_0$ will be small and they will compensate for large values of
$\varepsilon_i^{\star}$. Therefore, the iterative paraperspective algorithm is
likely to converge for those configurations where the weak perspective one
diverges. Notice that if the point $p_0$ thus computed lies in the vicinity of the optical
axis then the paraperspective and weak perspective models will yield the same results.

\section{The orthogonality constraint}
\label{section:The-orthogonality-constraint}

The algorithms described so far are linear and in the general case
(non coplanar object points) they don't guarantee
that the matrix describing the orientation of the object frame with
respect to the camera frame is a rotation matrix. This matrix is formed by
the three row vectors $\ivect ^T$, $\jvect ^T$, and $\kvect ^T$. These vectors
are updated at each step of the pose algorithms and we want to compute a
rotation matrix from these three vectors. In other words, we seek a rotation
matrix $R$ which verifies:
\[
R \left(
\begin{array}{c}
        \ivect ^T  \\
        \jvect ^T \\
        \kvect ^T 
    \end{array} \right) ^T
= I_{3\times 3}
\]
This expression can be written as:
\begin{eqnarray*}
R \ivect & = & \evect _1 \\
R \jvect & = & \evect _2 \\
R \kvect & = & \evect _3
\end{eqnarray*}
where $\evect_i$ is the $i^{th}$ column of the identity matrix.
The solution is given by the rotation matrix which minimizes the following
criterion:
\[
\min _{R} \sum _{i=1} ^3 \| R \vvect_i - \evect_i \|^2
\]
where $\vvect_i$ is either $\ivect$, $\jvect$, or $\kvect$. It is well known
that this minimization problem has a closed-form solution \cite{Faugeras93}.
Indeed, if the unknown rotation is represented by a unit quaternion $\qvect$,
then the minimization problem can be written as a quadratic form:
\[
\min _{\qvect}  (\qvect ^T B \qvect )
\]
where B is a 4$\times$4 symmetric, semi definite, and positive matrix.
Therefore, the quaternion $\qvect$ which minimizes this quadratic form is the
eigen vector associated with the smallest eigen value of $B$. One may 
notice that two vectors, i.e., $\ivect$ and $\jvect$ are sufficient for
computing the orthogonal matrix $R$.

\section{Sensitivity to camera calibration}
\label{section:Sensitivity-to-camera-calibration}

So far we assumed that the camera is calibrated, which means that the
intrinsic camera parameters are known, i.e., eqs. (\ref{eq:u-intrinsic}) and (\ref{eq:v-intrinsic}). It is well known that camera calibration is difficult and
that the camera parameters vary a lot. In this section we analyse the
sensitivity of the pose algorithms that we just described with respect to
variations of the intrinsic camera parameters.

We consider the equations associated with full
perspective. By combining eqs. (\ref{eq:u-intrinsic}) and (\ref{eq:v-intrinsic}) with eqs. (\ref{eq:x-dem-perspective}) and
(\ref{eq:y-dem-perspective}) we obtain:
\begin{eqnarray}
\label{eq:u-dem-persp}
u_i (1 + \varepsilon _i) - u_0 - u_c \varepsilon _i = \alpha _u \Ivect \cdot \Pvect
_i \\
\label{eq:v-dem-persp}
v_i (1 + \varepsilon _i) - v_0 - v_c \varepsilon _i = \alpha _v \Jvect \cdot \Pvect
_i 
\end{eqnarray}

When a camera is calibrated, relatively stable values may be obtained for
the horizontal and vertical scale factors $\alpha_v$ and $\alpha_u$ while the
image coordinates of the optical center, $u_c$ and $v_c$ can 
vary with up to 10\% of their
nominal values \cite{Tsai87}, \cite{Faugeras93}. 
By inspecting the previous equations it is easy to notice that
the influence of $u_c$ and $v_c$ is weighted by $\varepsilon_i$. For any object
point $P_i$, $\varepsilon_i$ is the projection of the vector $P_0P_i$ onto the
optical axis, divided by the z-component of the displacement vector $\tvect$.

If the object is too far from the camera, the influence of $u_c$ and $v_c$ can
be neglected but one needs to detect image points with great accuracy. If the
object is too close to the camera, the influence of $u_c$ and $v_c$ becomes
quite large. Therefore, in practice, the object should be at a distance from
the camera that is roughly 5 to 10 times the size of the object. In that case
the values of $\varepsilon_i$ will be between 0.1 and 0.2 and the error on
$u_c$ and $v_c$ will be scaled down by a factor of 5 to 10. These remarks are
validated by experimental results as reported in the next section.

\section{Experiments}
\label{section:Experiments}

In this section we study the performances of the iterative weak and para
perspective algorithms. Two types of performances are studied:
\begin{itemize}
\item the precision of pose as a function of position and orientation of the
object with respect to the camera in the presence of image and/or camera
noise, and
\item the convergence of the iterative pose algorithms as a function of
position and orientation of the object with respect to the camera.
\end{itemize}

In the first class of experiments (precision) we compare the results obtained
with three algorithms: the two linear algorithms described above and a
non-linear algorithm \cite{PhongHoraud95}. In the second class of experiments
(convergence) we compare the two linear algorithms. In both classes, the simulated 
object is a configuration of four points (tetrahedron), such that the three line segments 
joining the reference point to the other three points are equal and perpendicular to each other. 

For each experiment we fix a number of positions of the object with respect to
the camera and for each such position the object is rotated at $1000$ random 
orientations. These experiments are repeated for various levels of image and/or camera noise.
The rotation matrices defining these $1000$ orientations are computed from Euler angles 
chosen by a random number generator in the range $[0, 2\pi]$. The position of an object 
with respect to the camera is described by the translation vector from the center of
projection of the camera to the origin of the object frame. More quantitatively, we
compute the error between the theoretical pose and the pose computed by an
algorithm. For each position we plot the average of this error over all the $1000$ orientations.
The pose errors are: orientation error and position error. The orientation error is defined as 
the rotation angle in degrees required to align the coordinate system of the object in 
its computed orientation with the coordinate system of the object in its theoretical orientation. 
The position error is defined as the norm of the vector which represents the difference between the two 
translation vectors: the computed one and the theoretical one, divided by the norm of the second vector.  
The horizontal coordinates of all the following plots represents the z-component of 
the translation vector scaled by the object size.

\begin{figure}[htb]
\centering
\includegraphics[width=0.95\columnwidth]{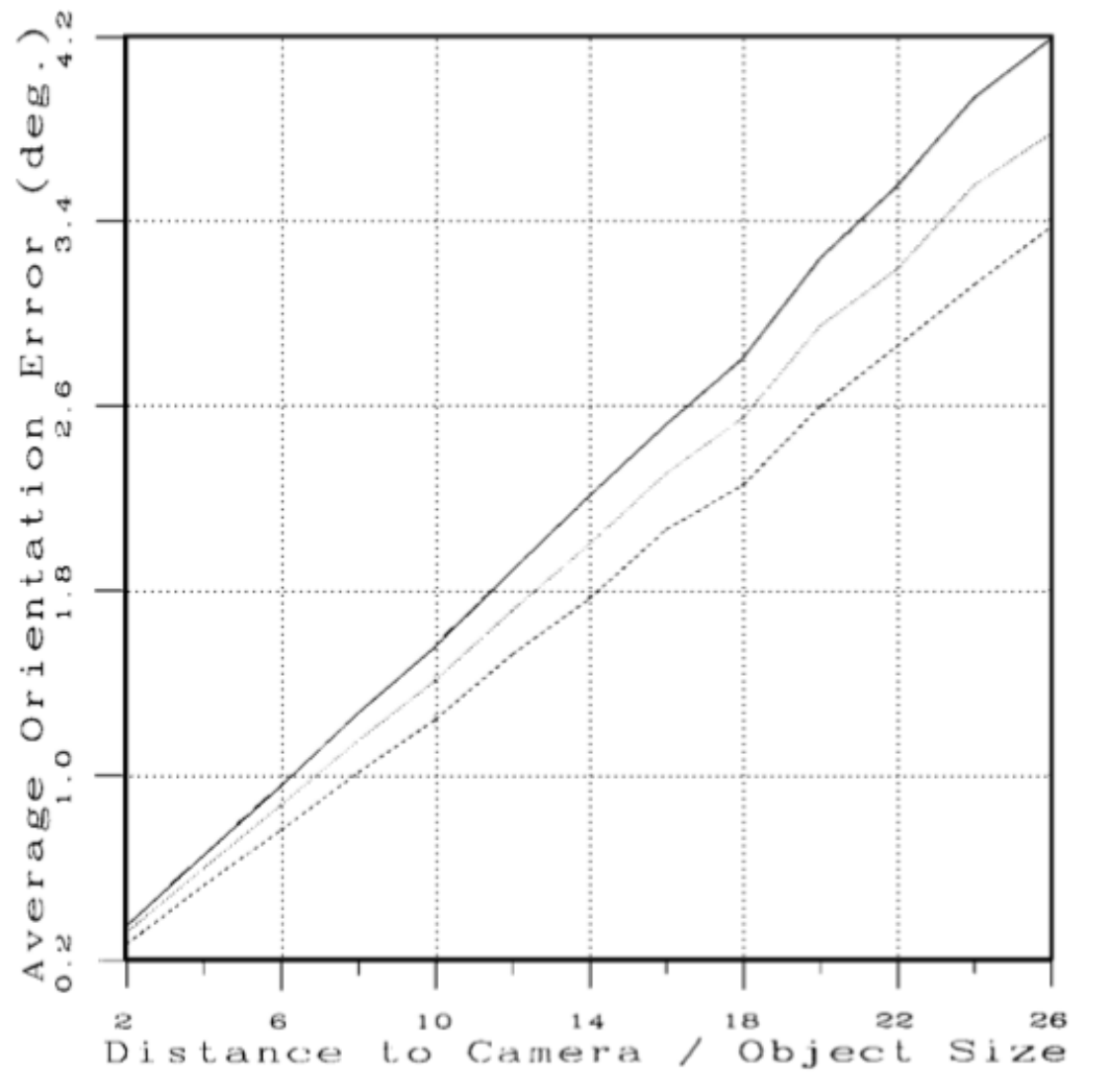}
\caption{Error in orientation as a function of depth in the presence
of image gaussian noise. The solid line corresponds to the iterative
paraperspective algorithm, the dotted line corresponds to the iterative
paraperspective algorithm combined with the orthogonal constraint, and
the dashed line corresponds to a non linear algorithm.}
\label{fig:average-orientation}
\end{figure}

\begin{figure}[htb]
\centering
\includegraphics[width=0.95\columnwidth]{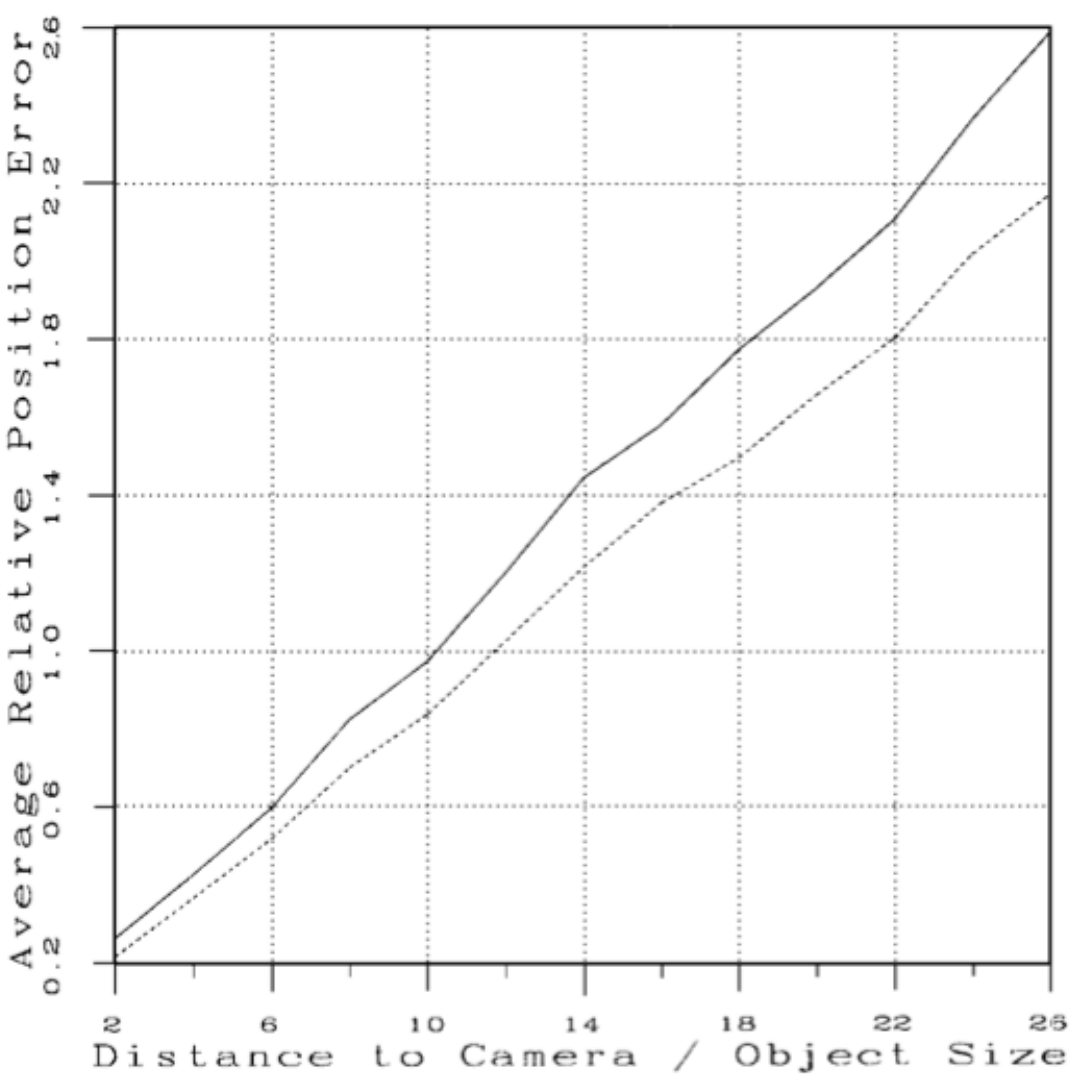}
\caption{Error in position as a function of depth in the presence
of image gaussian noise. The iterative weak and paraperspective algorithms
have the same behaviour (solid line) while the non linear algorithm (dashed 
line) performs slightly better.}
\label{fig:average-position}
\end{figure}

Figure \ref{fig:average-orientation} shows the variation of the error in
orientation in the presence of image gaussian noise as a function of the
ratio between the distance to the camera and the object size, where the
object size has been fixed to a constant value. The three curves in this
figure correspond to the iterative paraperspective algorithm (solid line),
the iterative paraperspective algorithm with the orthogonal constraint
described in Section~\ref{section:The-orthogonality-constraint} 
(dotted line) and to the non-linear algorithm
described in \cite{PhongHoraud95} (dashed). The iterative weak perspective
algorithm yields a curve that is identical to the paraperspective one. 
Figure \ref{fig:average-position} shows the variation of the relative error in
position in the presence of image gaussian noise as a function of the
ratio between the distance to the camera and the object size. The solid line
corresponds to the iterative weak and paraperspective algorithms (identical
behaviour) and the dashed line corresponds to the non-linear algorithm. In
both cases (Figure~\ref{fig:average-orientation} and 
Figure~\ref{fig:average-position}) the image data has been perturbed with
gaussian noise with a standard deviation equal to 1 pixel.

\begin{figure}[tb]
\centering
\includegraphics[width=0.95\columnwidth]{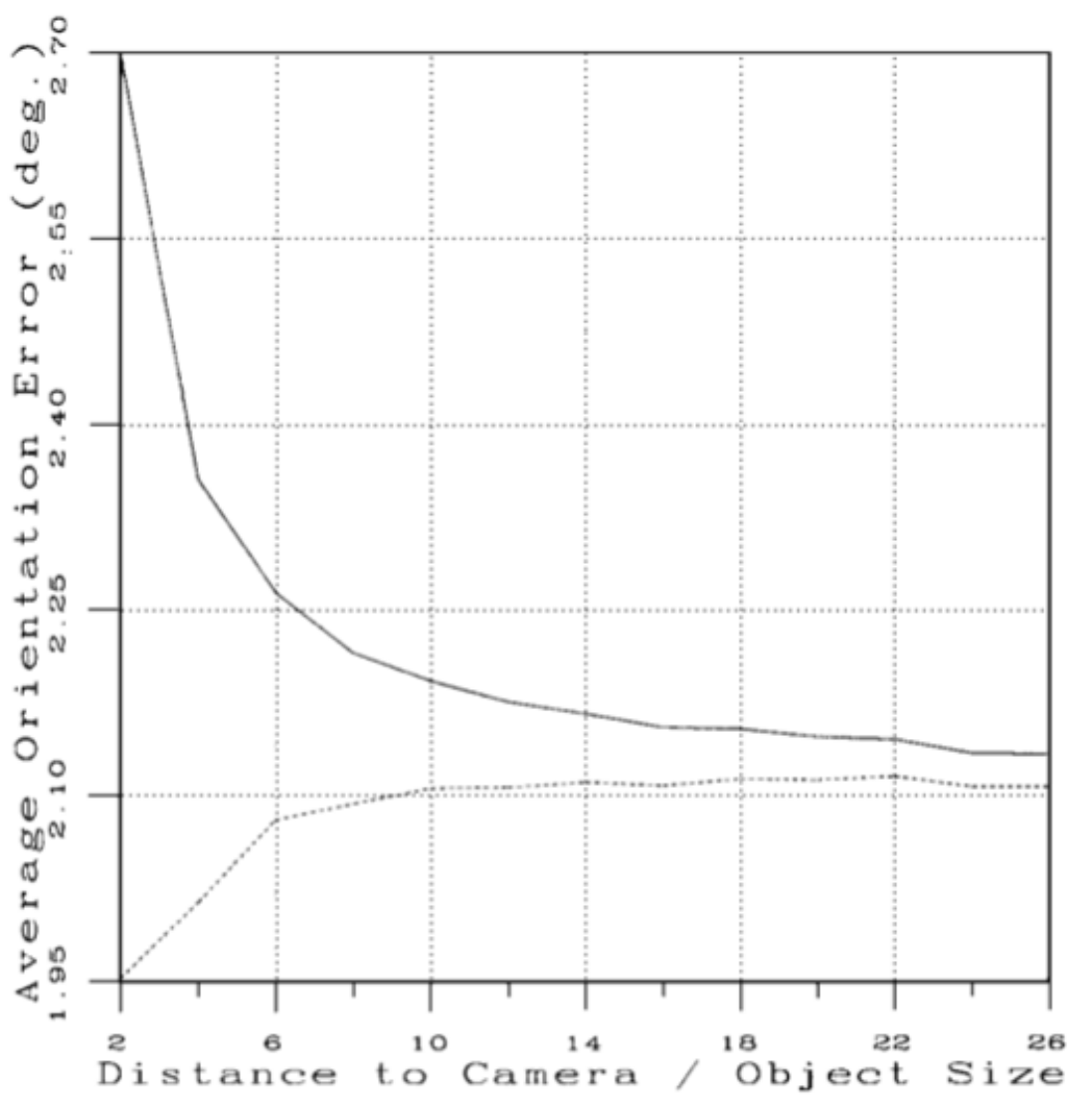}
\caption{Error in orientation as a function of depth in the presence
of camera uniform noise. The behaviour of the iterative paraperspective 
algorithm (solid line) is compared with a
non linear algorithm (dashed line).}
\label{fig:orientation-intrinsic}
\end{figure}

Figure~\ref{fig:orientation-intrinsic} shows the variation of the error in
orientation as a function of the distance to the camera scaled by the object
size. The difference between this figure and
Figure~\ref{fig:average-orientation} is that here we added uniformly
distributed noise to the intrinsic camera parameters. Namely the coordinates
of the optical axis in the image frame were perturbed with uniform noise in
the interval $[-15\%,\; +15\%]$. The behaviour of the iterative
paraperspective algorithm (solid line) is the one predicted by 
equations~(\ref{eq:u-dem-persp}) and (\ref{eq:v-dem-persp}) -- the effect of
an inacurate center of projection vanishes as the object is 
farther away from the camera. For short distances 
the non-linear algorithm (dashed line) is more
acurate than the iterative paraperspective algorithm.

\begin{figure}[tb]
\centering
\includegraphics[width=0.95\columnwidth]{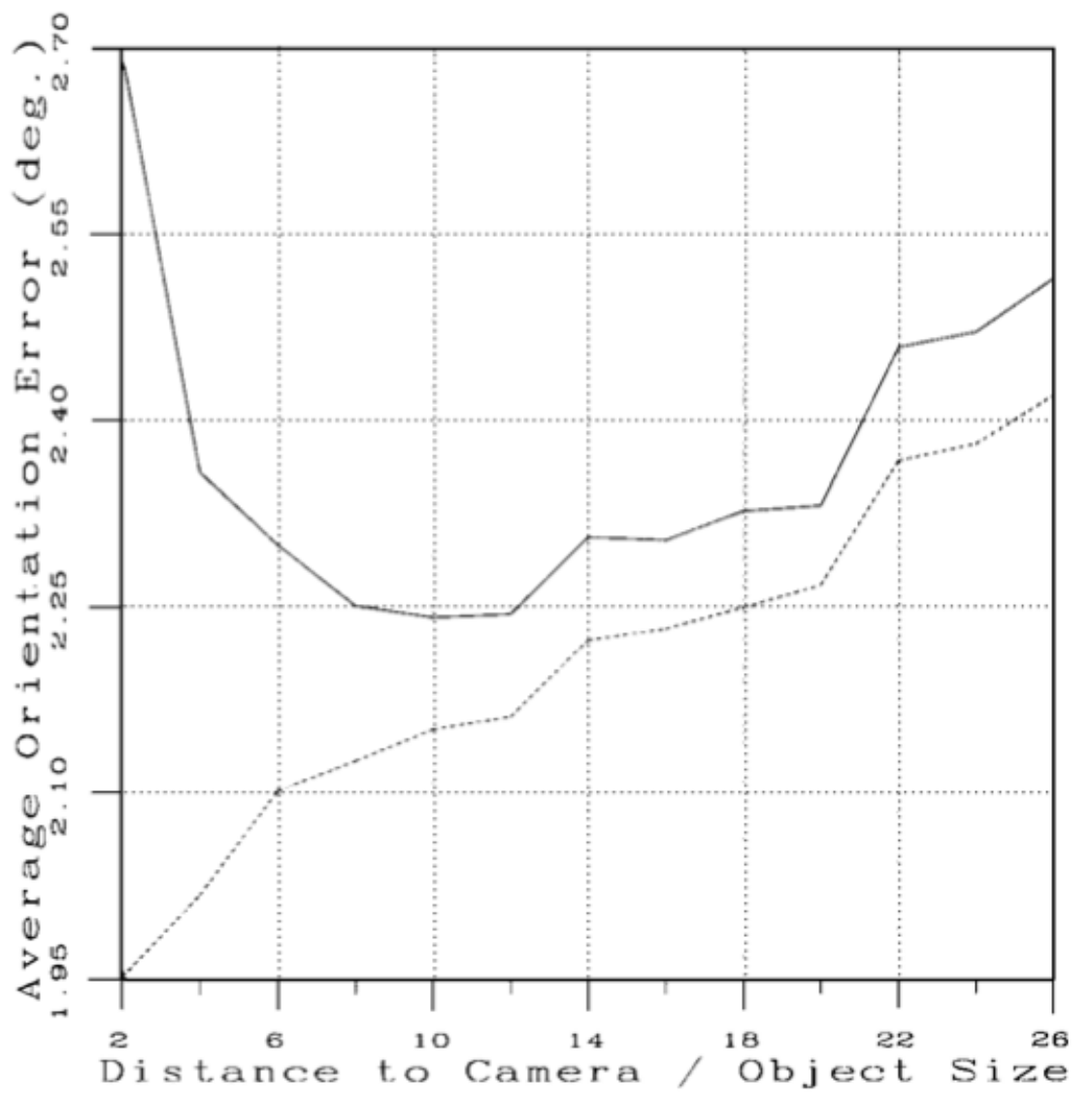}
\caption{Error in orientation as a function of depth in the presence
of camera and image uniform noise. The iterative paraperspective algorithm
(solid line) has a behaviour that differs from the non-linear algorithm
(dashed line).}
\label{fig:orientation-2noises}
\end{figure}

If we combine camera noise and image noise, then
we obtain two other curves shown on Figure~\ref{fig:orientation-2noises}.
Uniform pixel noise in the
interval 
$[-0.5,\; +0.5]$ has been added to the data. The iterative paraperspective
algorithm (solid line) has a minimum for a distance/size ratio equal to 10.
This is consistent with the theoretical predictions of Section~\ref{section:Sensitivity-to-camera-calibration}. We conclude that the optimal
behaviour of the iterative pose algorithms occur when the distance between the
object and the camera is 5 to 10 times the size of the object. In the presence
of both camera noise and image noise, the non linear
algorithm (dashed line) performs better than the linear algorithm. 

We compare now more thoroughly
the two iterative pose algorithms. More precisely we study
the convergence of these algorithms. We performed the following kinds of experiments.
The first kind is meant to compare the number of iterations needed by
each one of these algorithms to converge to the theoretical solution. In order 
to take into account the effect of the offset from the optical axis, we constrain the origin 
of the object frame to belong to a fixed line of sight. The object offset 
can now be defined as the angle between this line of sight and the optical axis.
For each such offset and for each position (depth) we randomly selected $1000$ 
different orientations and we run both algorithms for each such position and orientation. We plot the average value of the number of iterations over all the $1000$ random orientations.

\begin{figure}[htb]
\centering
\includegraphics[width=0.92\columnwidth]{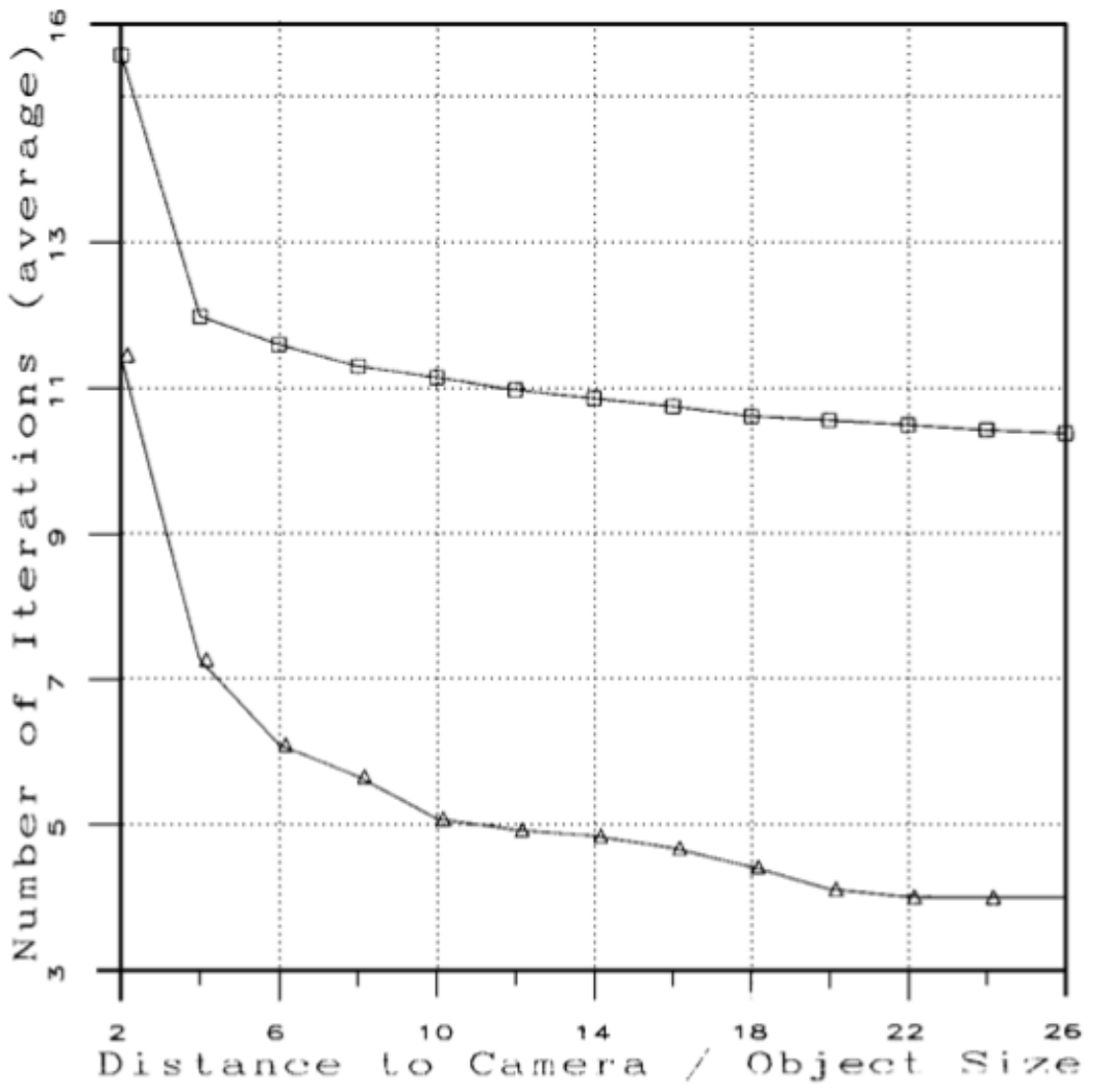}
\caption{Speed of convergence as a function of depth, the offset is equal to $23^o$ (triangles: paraperspective, squares: weak perspective).}
\label{fig:convergence-iter}
\end{figure}

\begin{figure}[htb]
\centering
\includegraphics[width=0.95\columnwidth]{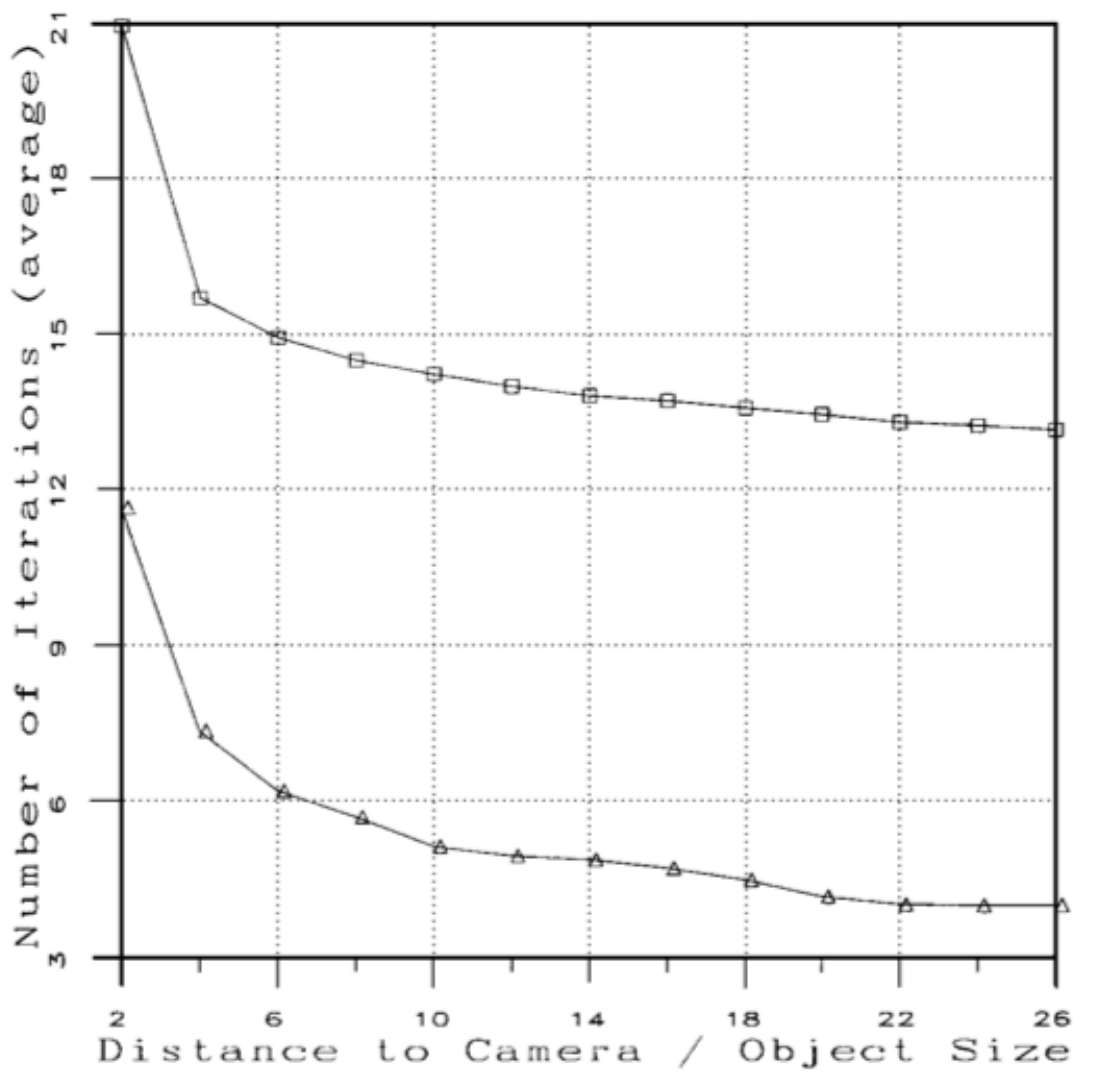}
\caption{Speed of convergence as a function of depth, the offset is equal to $30^o$ (triangles: paraperspective, squares: weak perspective).}
\label{fig:convergence-iter-fadi}
\end{figure}

Figures~\ref{fig:convergence-iter} and~\ref{fig:convergence-iter-fadi} 
show the number 
of iterations as a function of depth - more precisely, the distance from the object 
to the camera scaled by the object size. The lines with squares correspond to the 
weak perspective algorithm, the lines with triangles correspond to the paraperspective
 algorithm. These figures correspond respectively to two offsets, $23^o$ and 
$30^o$. On an average, the paraperspective algorithm converges 2 to 3 times faster 
than the weak perspective algorithm.

\begin{figure}[htb]
\centering
\includegraphics[width=0.95\columnwidth]{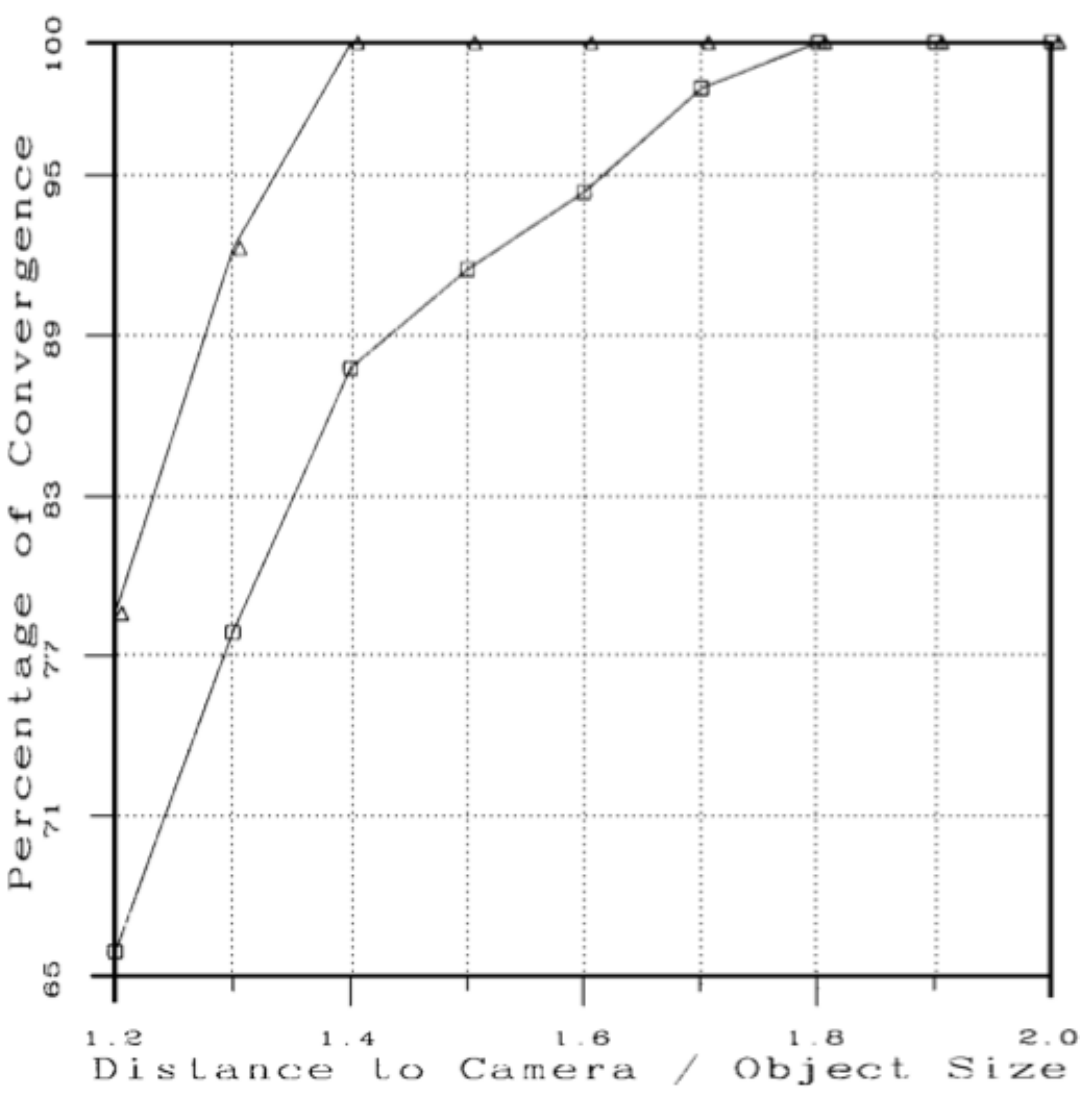}
\caption{Rate of convergence as a function of depth, the offset is equal to $30^o$ (triangles: paraperspective, squares: weak perspective).}
\label{fig:convergence-success-fadi}
\end{figure}

\begin{figure}[htb]
\centering
\includegraphics[width=0.95\columnwidth]{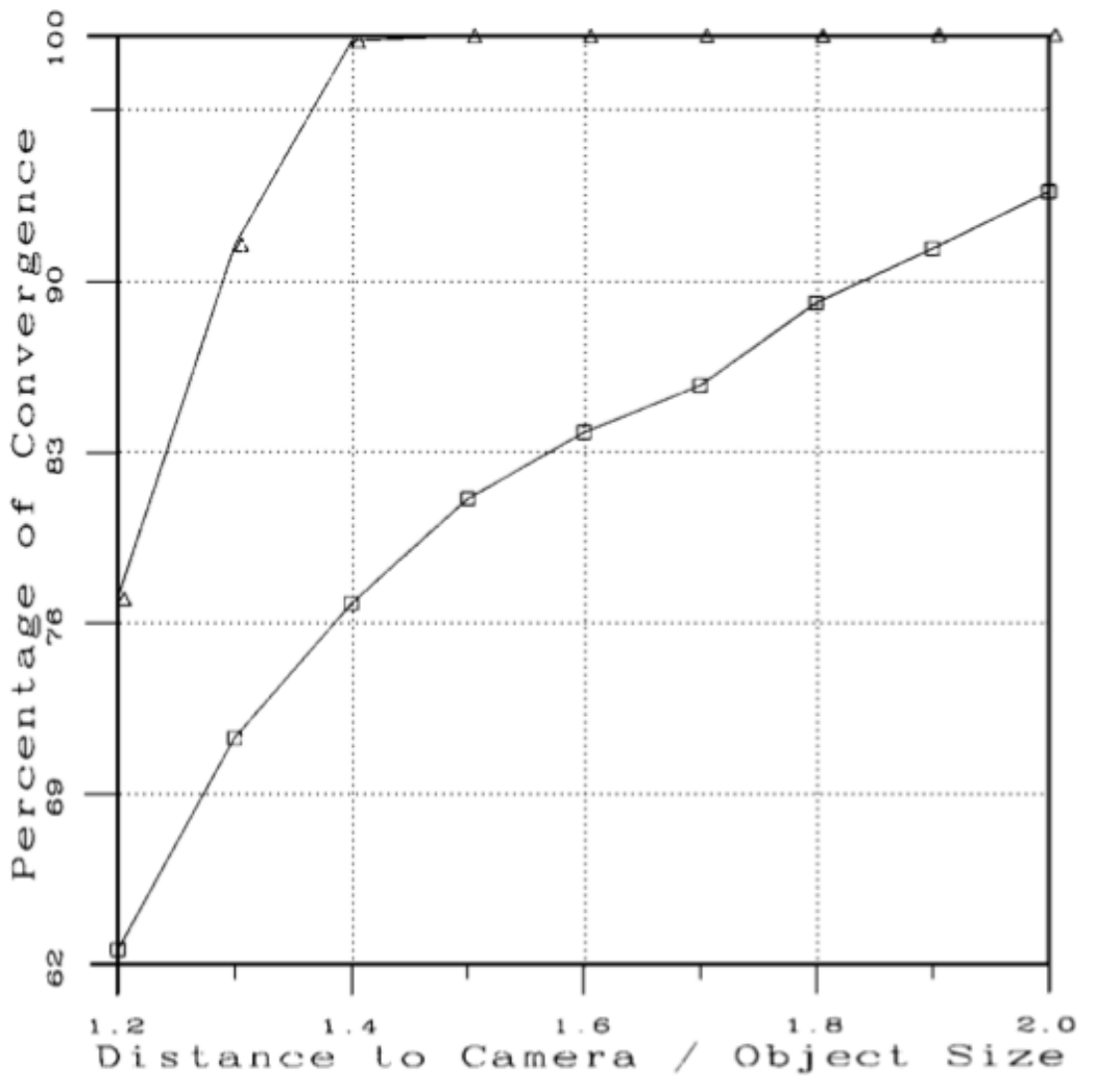}
\caption{Rate of convergence as a function of depth, the offset is equal to $35^o$ (triangles: paraperspective, squares: weak perspective).}
\label{fig:convergence-success}
\end{figure}

The second kind is meant to compare the rate of convergence of each one
of these algorithms for small distance/size ratios. 
Figures~\ref{fig:convergence-success-fadi} and~\ref{fig:convergence-success} show 
the percentage of the convergence of both algorithms as 
a function of depth. These figures correspond respectively to two offsets, $30^o$ and 
$35^o$.  As studied above in the paper, 
the weak perspective algorithm has an optimal behaviour when the object is either 
quite far from the camera or close to the optical axis. In theory, the paraperspective algorithm does not suffer from the offset limitations (see figures~\ref{fig:convergence-iter}, \ref{fig:convergence-iter-fadi},
\ref{fig:convergence-success-fadi} and \ref{fig:convergence-success}). For example (Figure~\ref{fig:convergence-success}), when the 
distance from the object to the camera is 1.4 times the size of the object,
the paraperspective algorithm has always converged (solid line  with
triangles) while the weak perspective algorithm has converged only in 76 \% of
the configurations. 


\section{Examples}
\label{section:Examples}

\begin{figure}[htb]
\centering
\frame{\includegraphics[width=0.49\columnwidth]{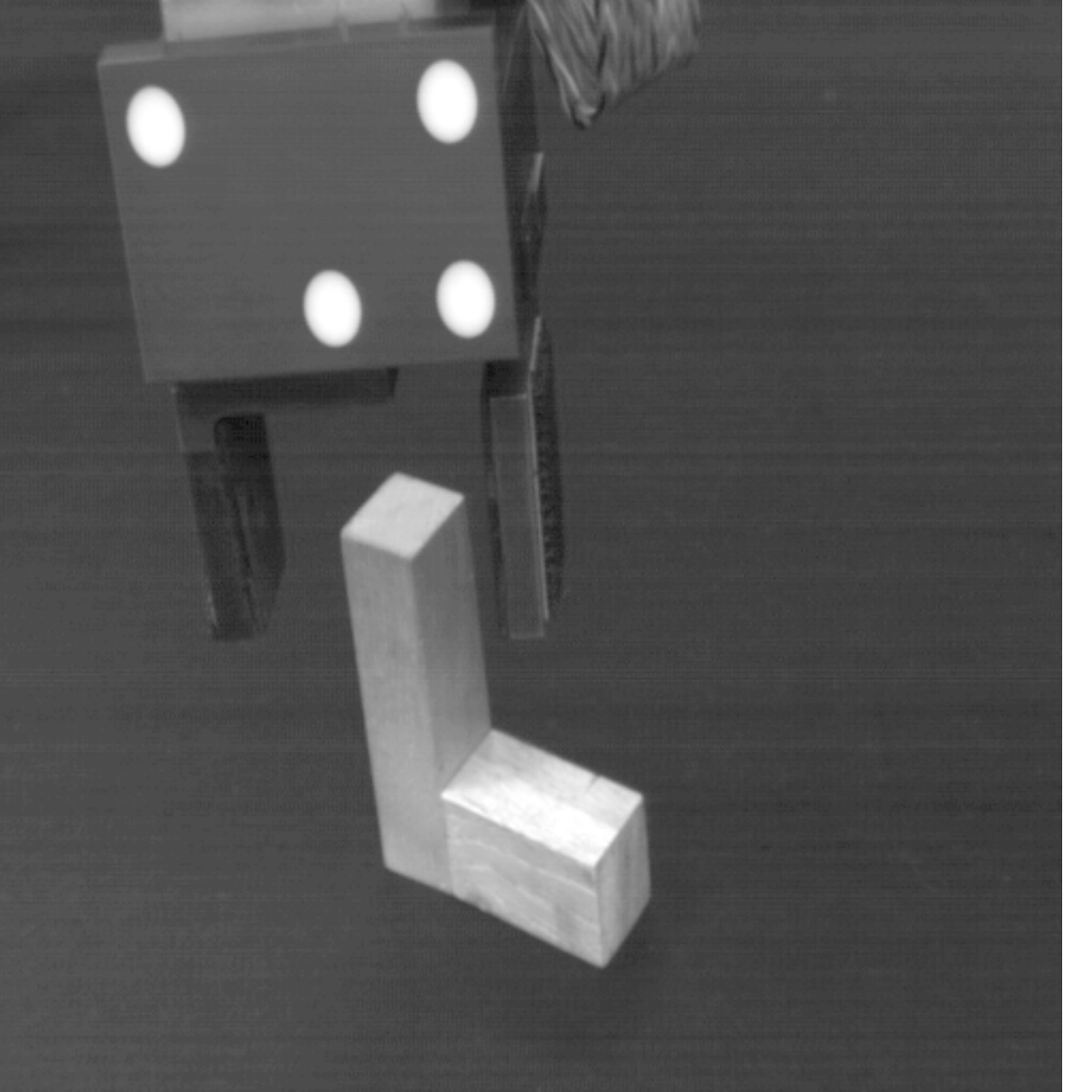}}
\frame{\includegraphics[width=0.49\columnwidth]{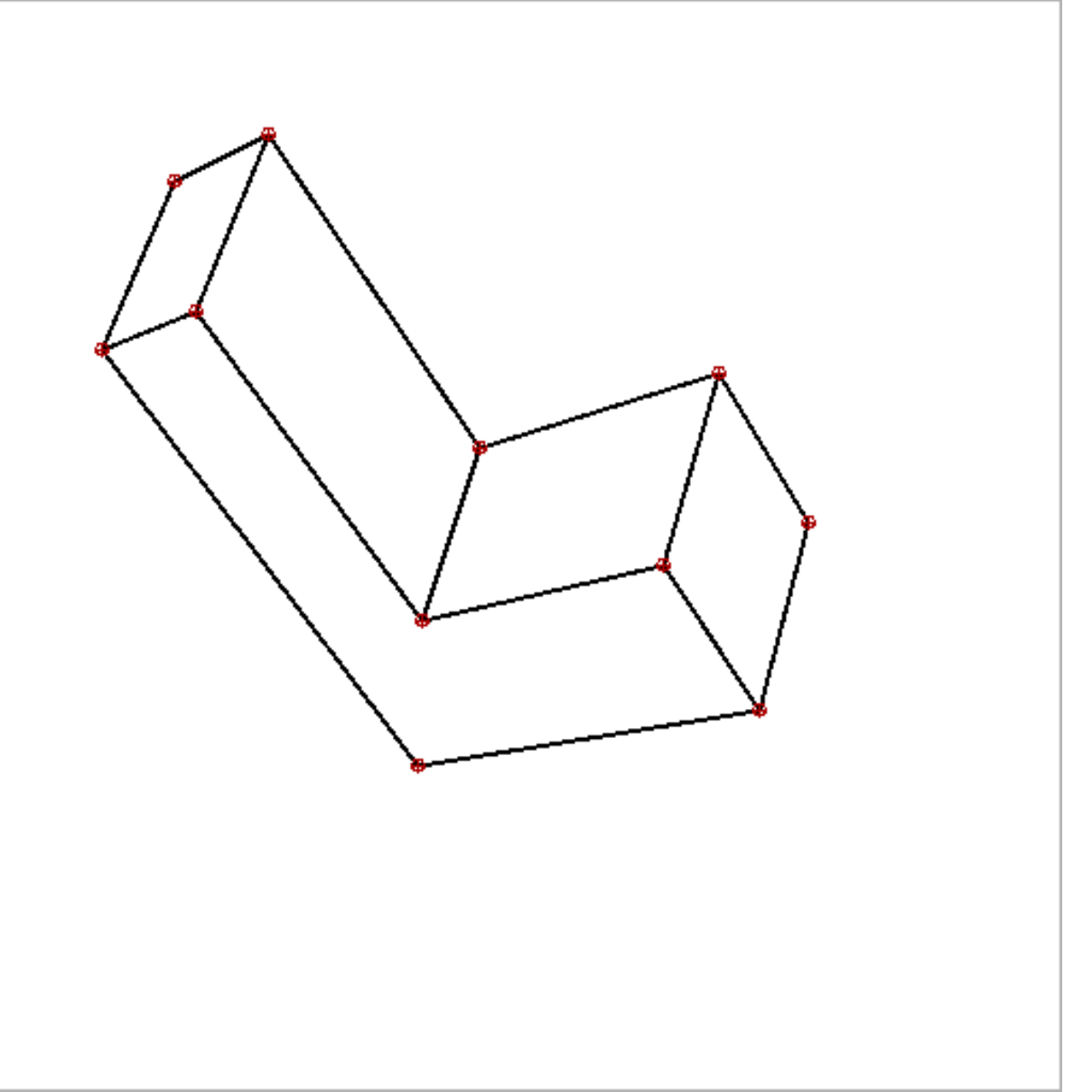}}
\frame{\includegraphics[width=0.49\columnwidth]{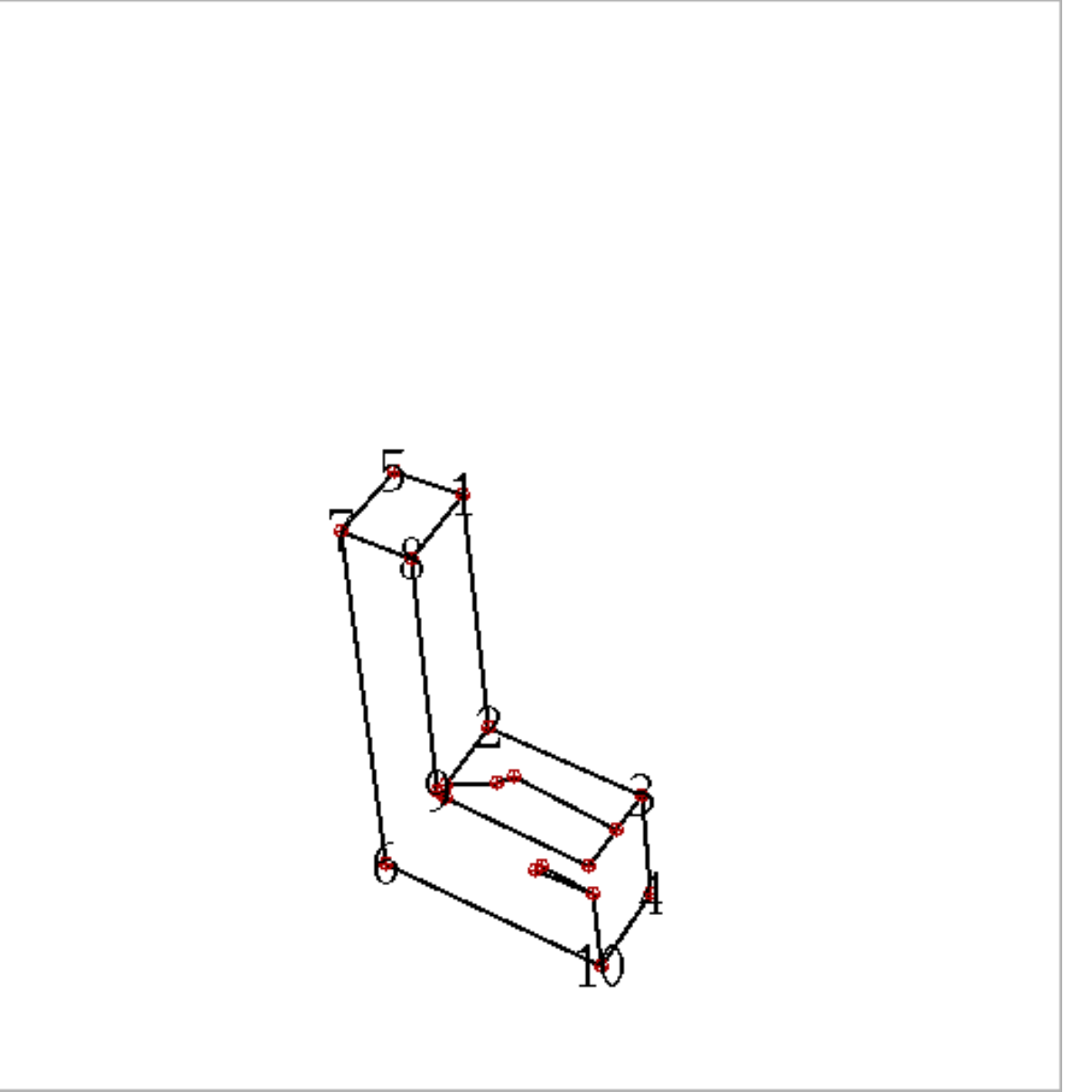}}
\frame{\includegraphics[width=0.49\columnwidth]{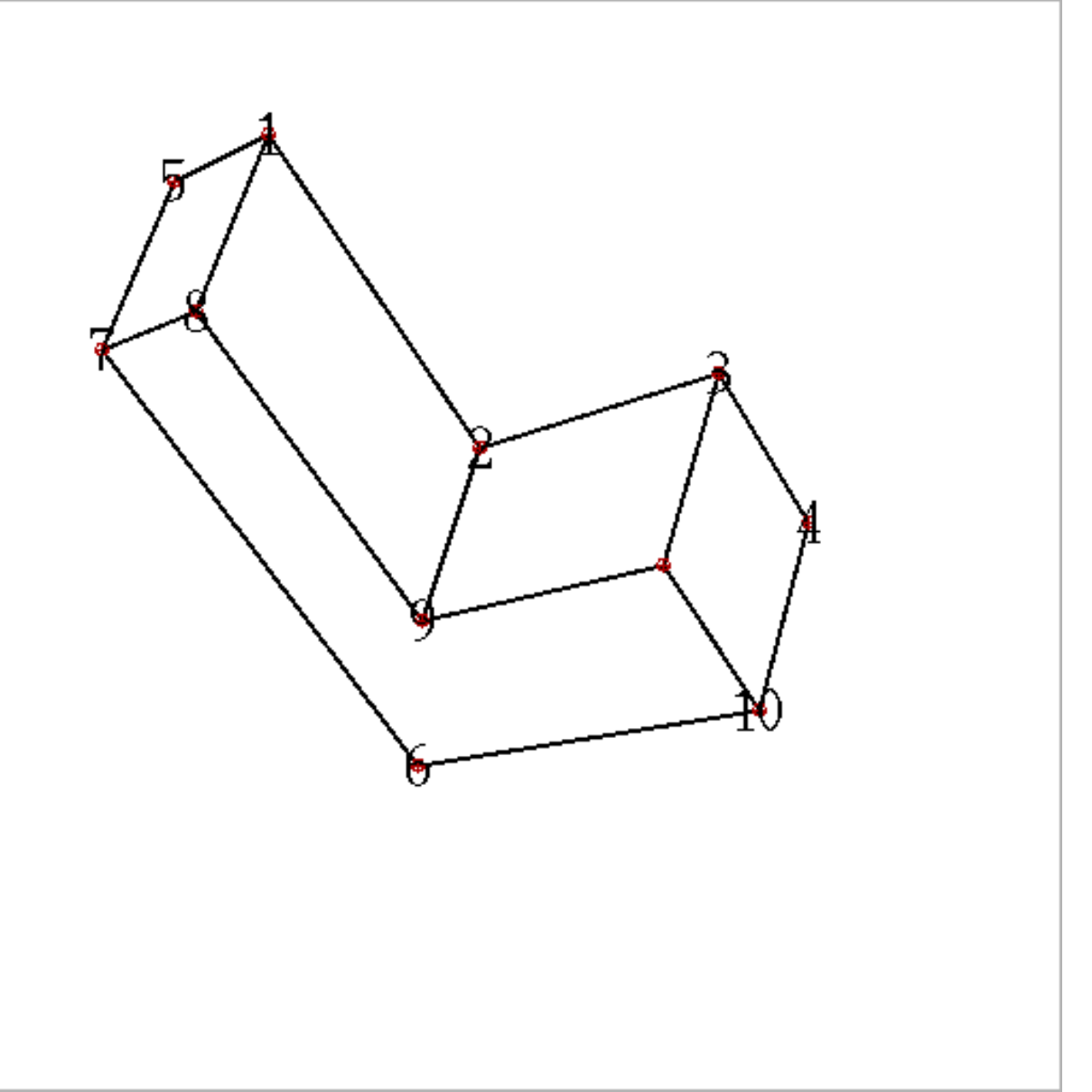}}
\frame{\includegraphics[width=0.49\columnwidth]{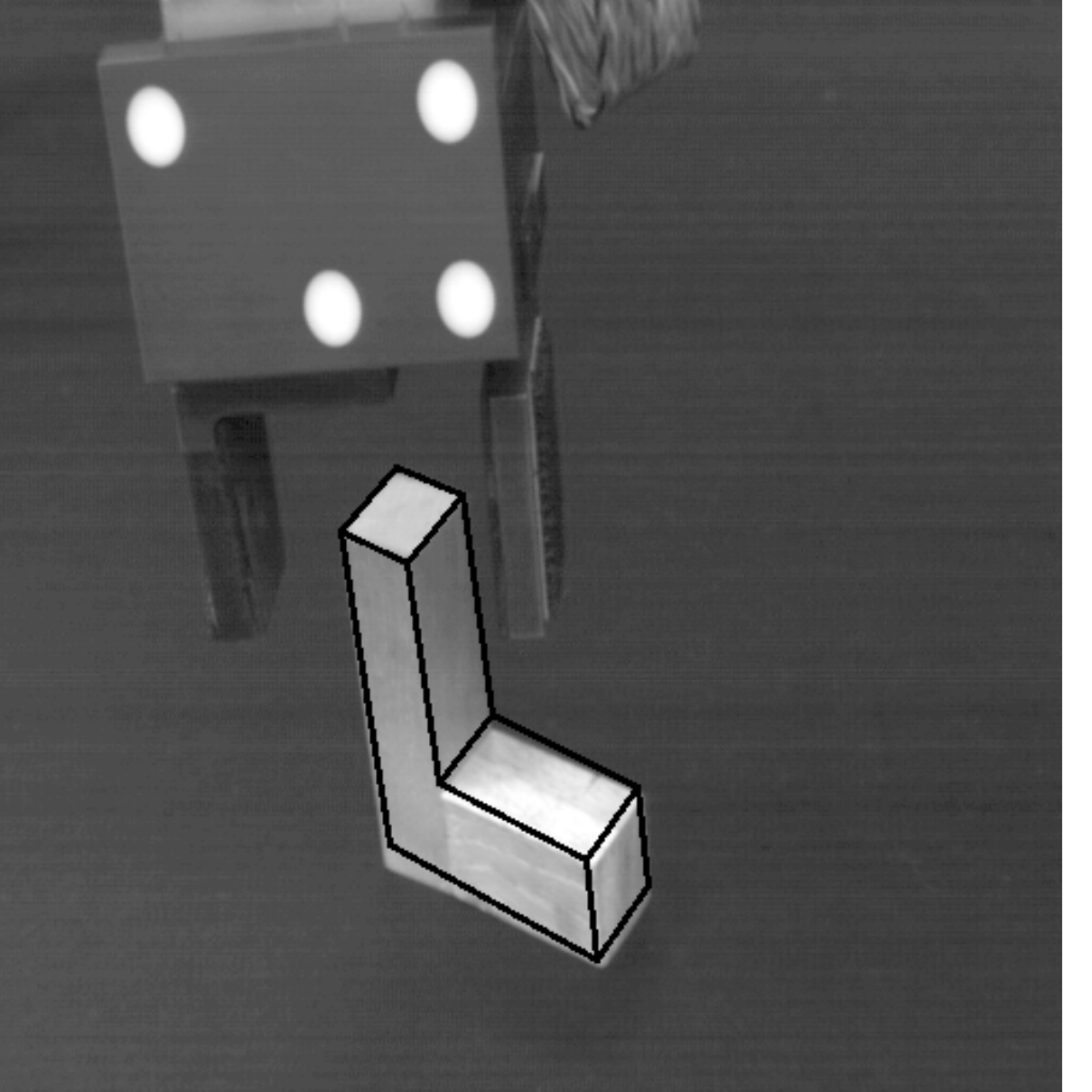}}
\frame{\includegraphics[width=0.49\columnwidth]{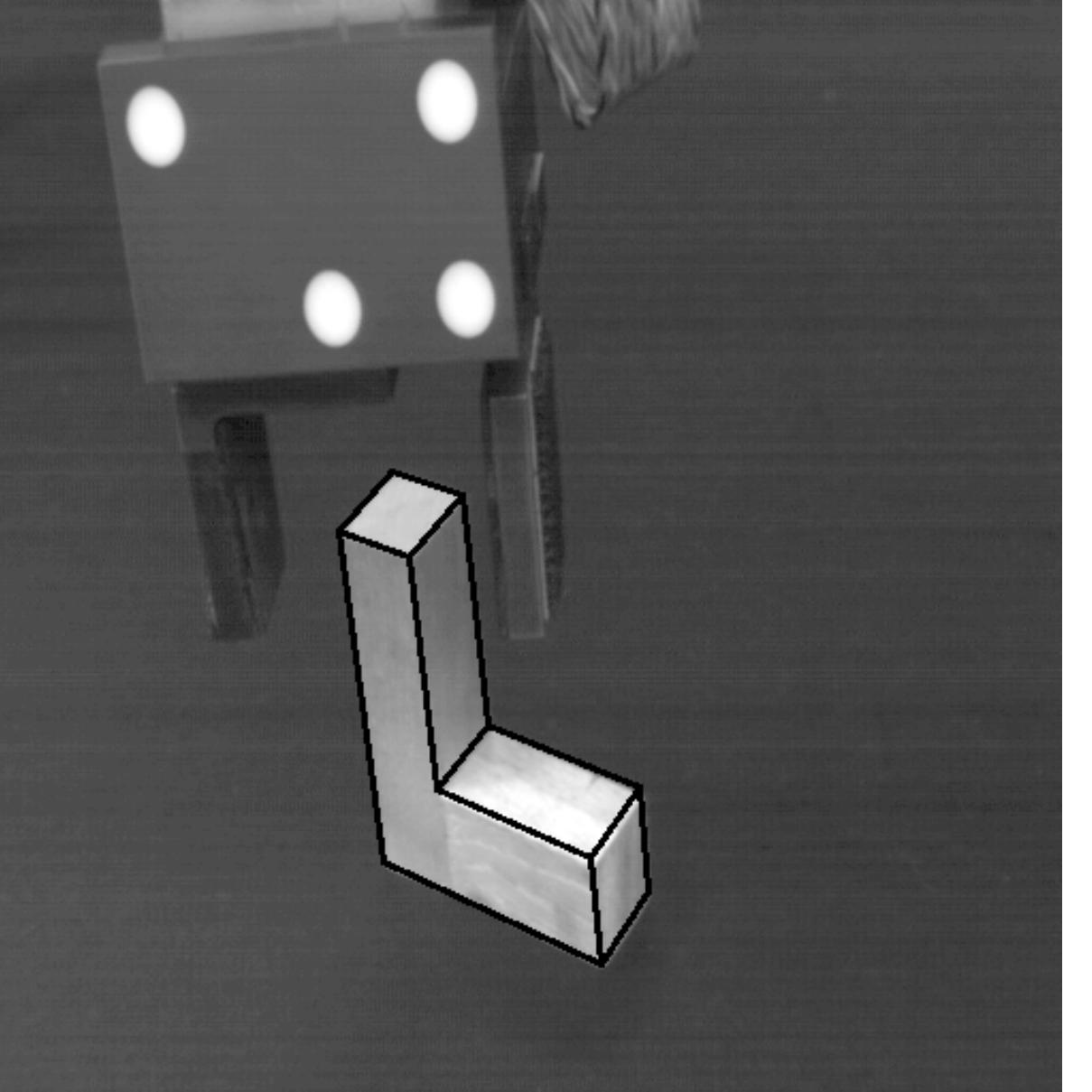}}
\caption{An example of applying the iterative paraperspective algorithm to a
non coplanar set of point correspondences.}
\label{fig:first-example}
\end{figure}

The iterative algorithms proposed in this paper have been applied to a number
of images. For example, Figure~\ref{fig:first-example} shows the image of a
polyhedral object to be located (top-left) and its wireframe representation
(top-right) whose position and orientation with respect to the camera has been
roughly hypothesized. Both the image and the model are described by a network
of straight lines and junctions which are matched using a method 
similar to the one described in \cite{Gros93d}. Using this technique,
10 junctions were correctly matched (middle). The first iteration of the
algorithm found a ``paraperspective pose" (bottom-left). After only
three iterations the algorithm correctly determined the position and
orientation of the polyhedral object with respect to the camera
(bottom-right).

\begin{figure}[htb]
\centering
\frame{\includegraphics[width=0.49\columnwidth]{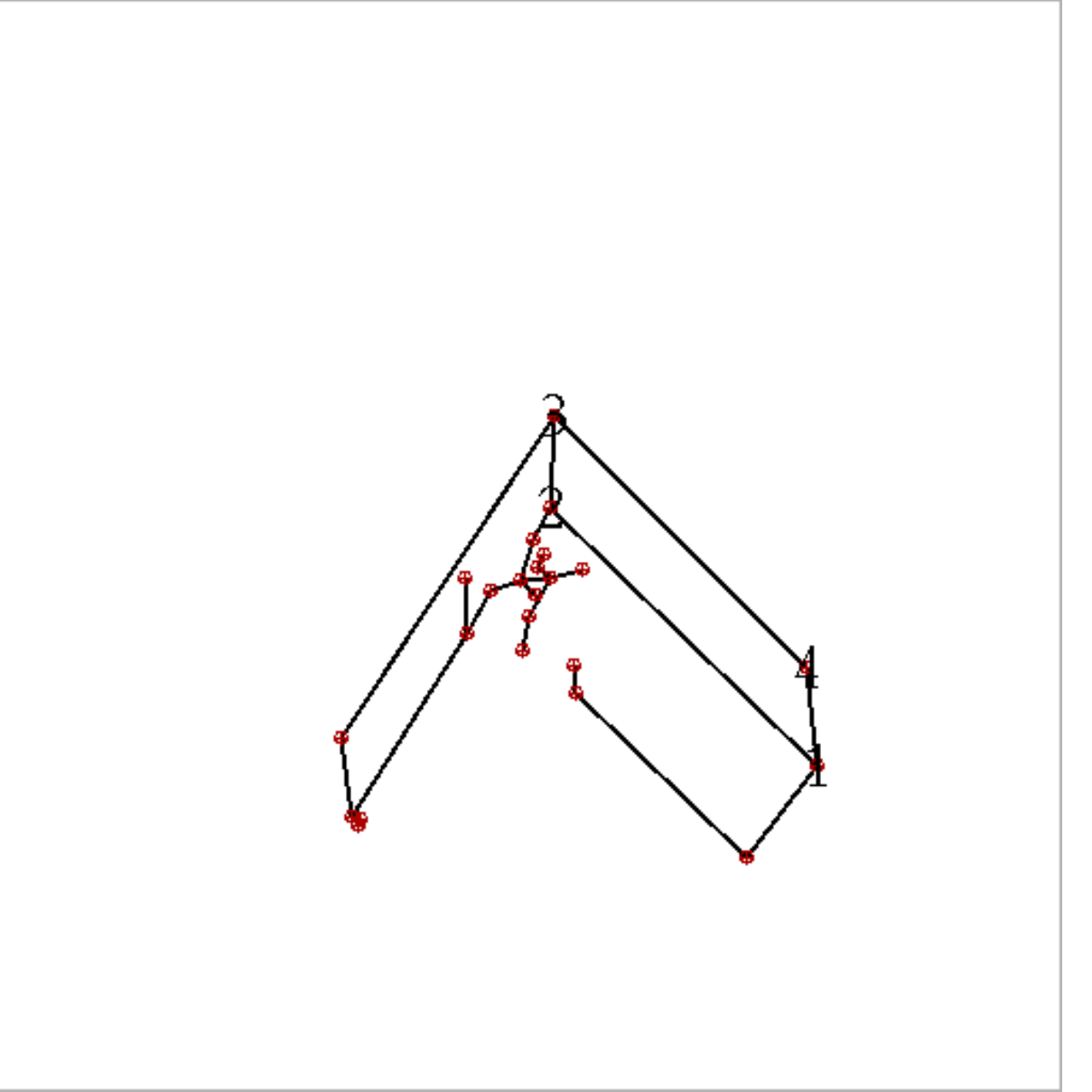}}
\frame{\includegraphics[width=0.49\columnwidth]{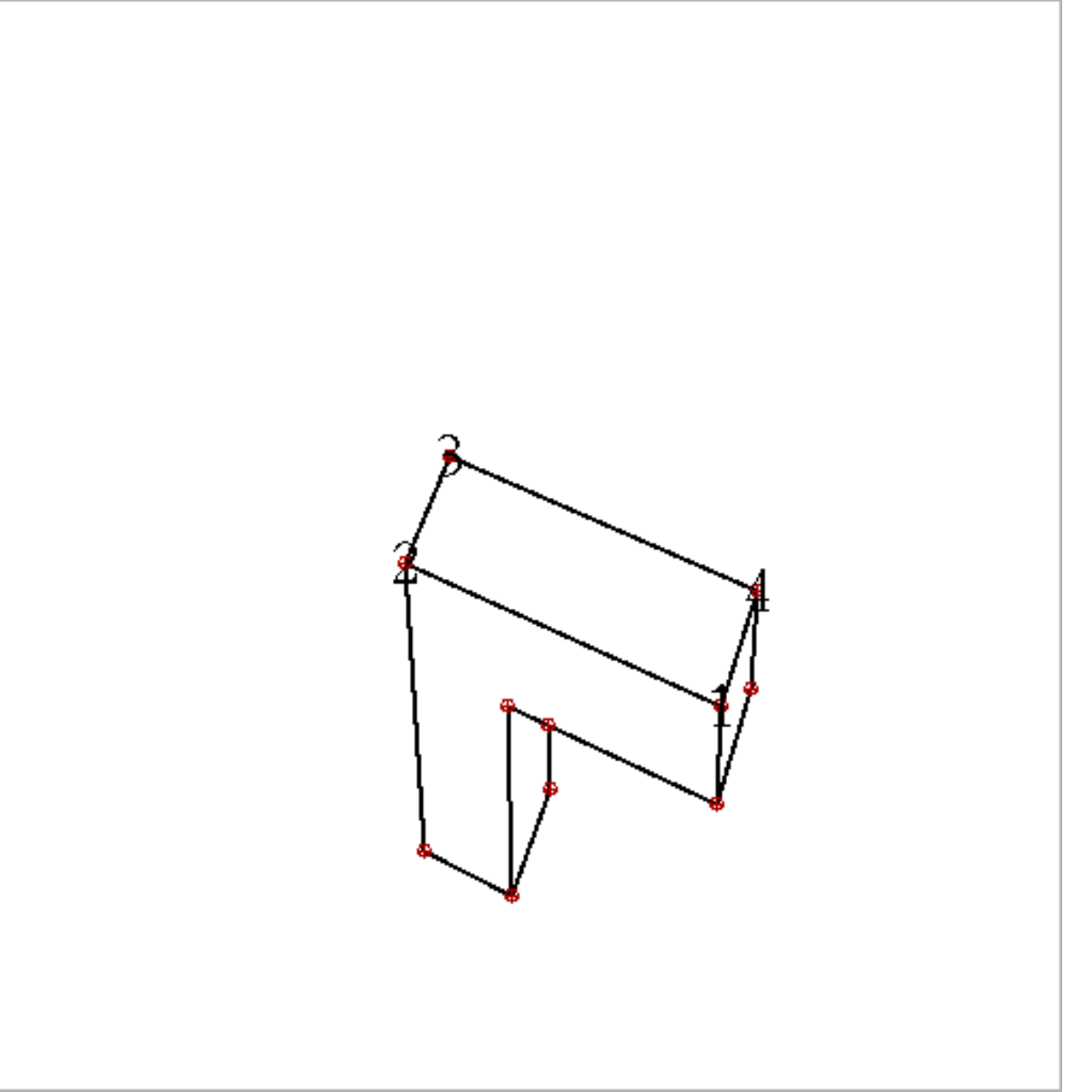}}
\frame{\includegraphics[width=0.49\columnwidth]{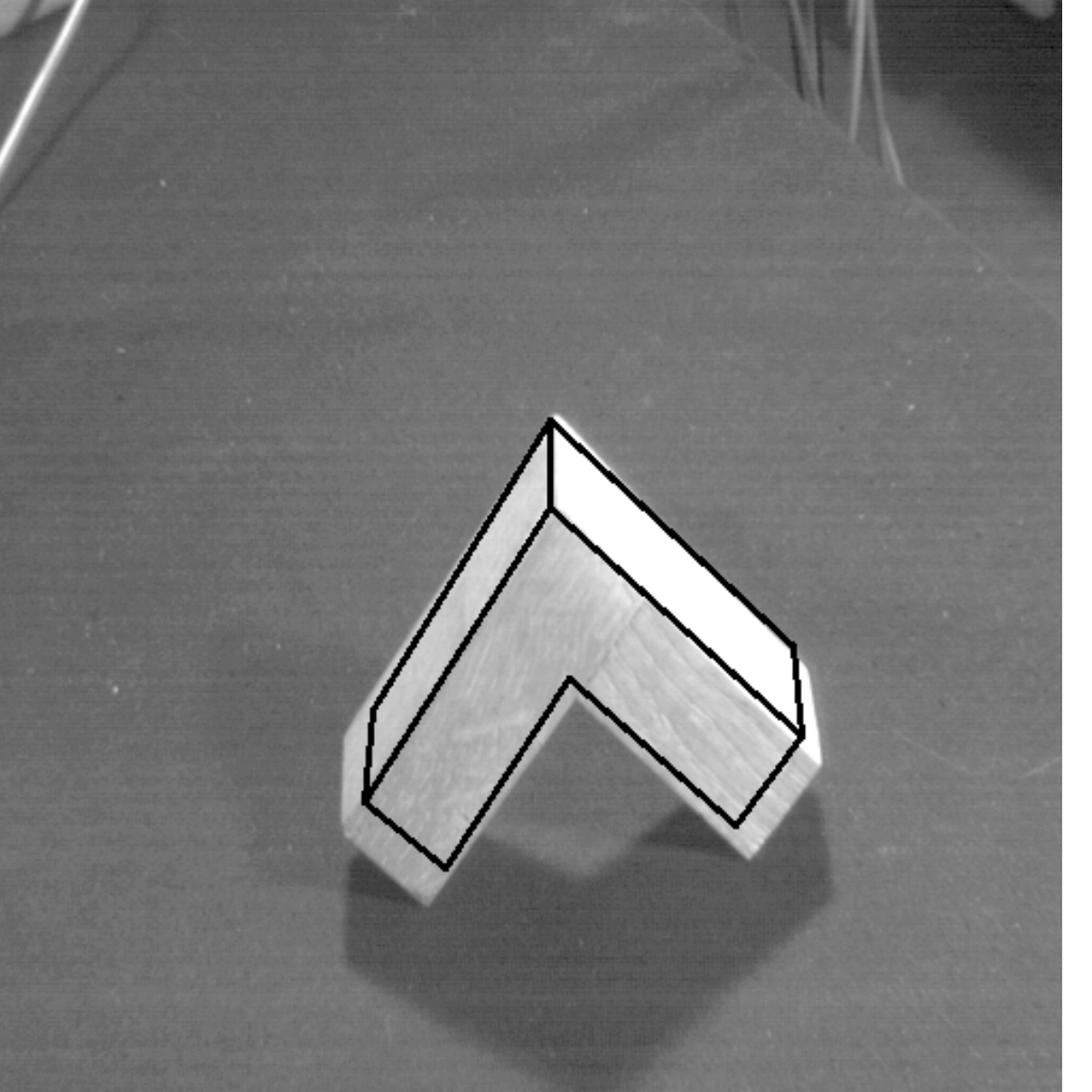}}
\frame{\includegraphics[width=0.49\columnwidth]{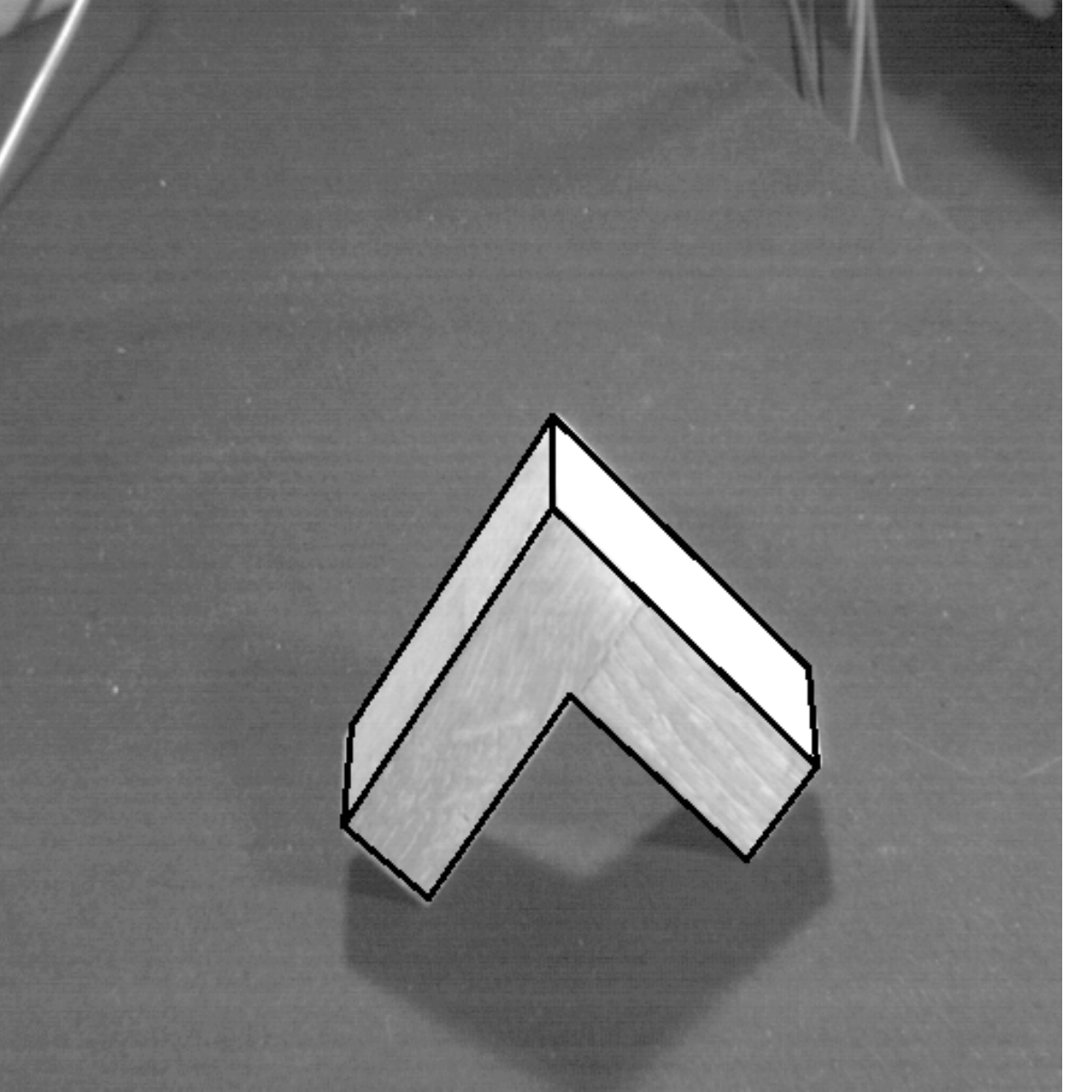}}
\caption{An example of applying the iterative paraperspective algorithm to a
set of 4 coplanar point correspondences.}
\label{fig:second-example}
\end{figure}

The second example, i.e., Figure~\ref{fig:second-example} illustrates a
situation where only 4 junctions have been matched corresponding to coplanar
model vertices (top). The first iteration found an approximated pose
(bottom-left) and after 3 iterations the algortihm converged to the
correct pose (bottom-right). 

In both examples the iterative paraperspective algorithm converged after
$0.7\times 10^{-3}$ seconds on a Sun/Sparc10.

\begin{figure}[htbp]
\centering
\includegraphics[width=0.99\columnwidth]{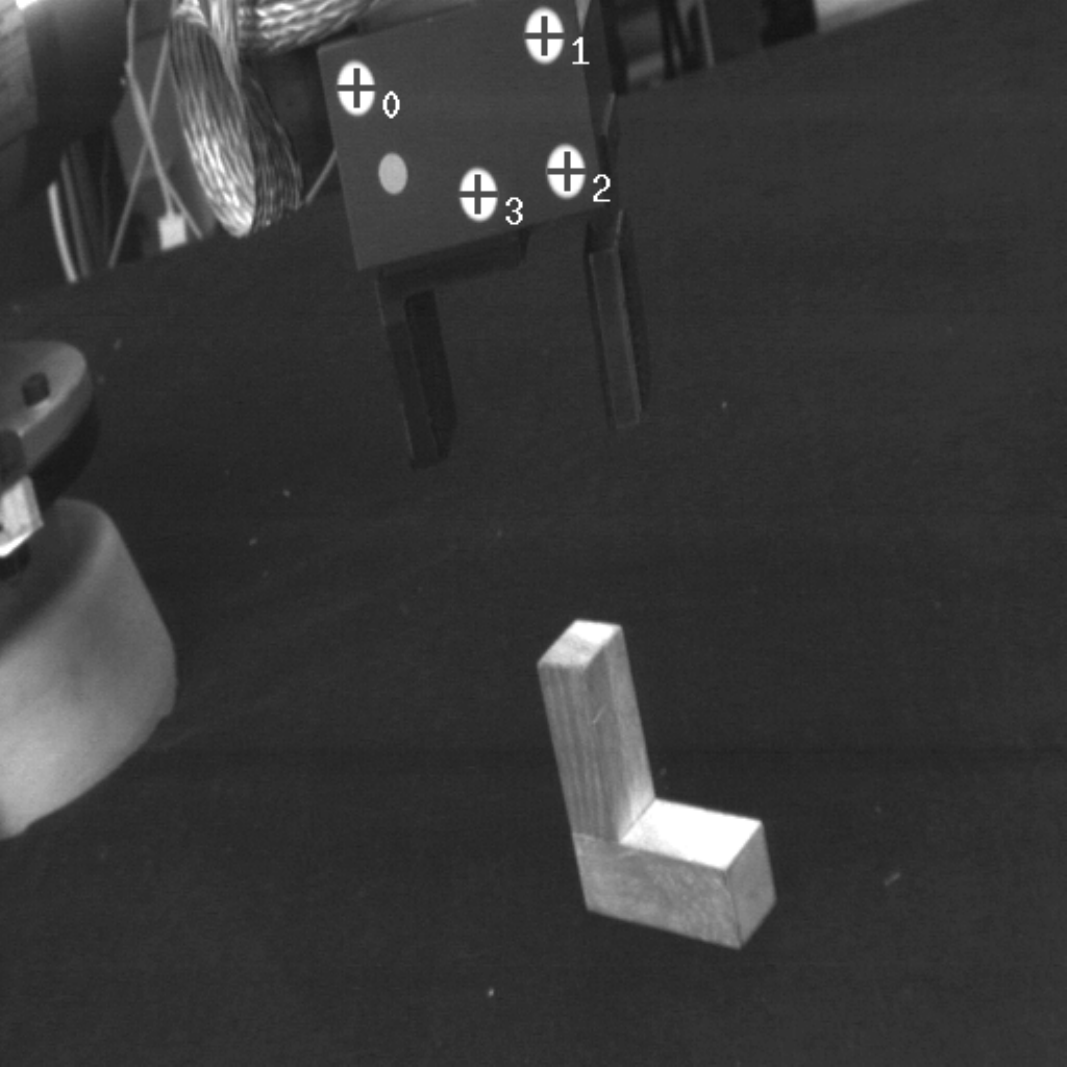}
\caption{The four image dots correspond to a coplanar set of object points.
Because the object is farther away from the optical axis, the iterative
paraperspective method performs better in this case than the iterative weak
perspective method.}
\label{fig:pose-plaque}
\end{figure}

The third example illustrates the behaviour of the linear and non linear
algorithm with 4 coplanar object points, e.g., Figure~\ref{fig:pose-plaque}
and Table~\ref{table:pose-plaque}. As one may easily notice, the
iterative paraperspective algorithm is faster (both in term of number of
iterations and CPU time) than the iterative weak perspective algorithm. The
non-linear algorithm was initialized with the solution obtained by using a paraperspective 
model, i.e., the pose computed at the first iteration of the iterative paraperspective 
algorithm.

\begin{table}[htbp]
\begin{center}
\begin{tabular}{||c|c|c||}
 \hline
Method 		&  Number of iterations		& CPU time (ms) \\ \hline
It. weak persp. &	10			& 4.3 \\ \hline
It. parapersp.  &	5			& 2.9 \\ \hline
Non linear	&	91			& 65.3 \\ \hline 
\end{tabular}
\end{center}
\caption{A comparison of the performance of the three algorithms.
The computer being used is a Sun Sparc10/51.}
\label{table:pose-plaque}
\end{table}

\begin{figure}[htbp]
\centering
\includegraphics[width=0.49\columnwidth]{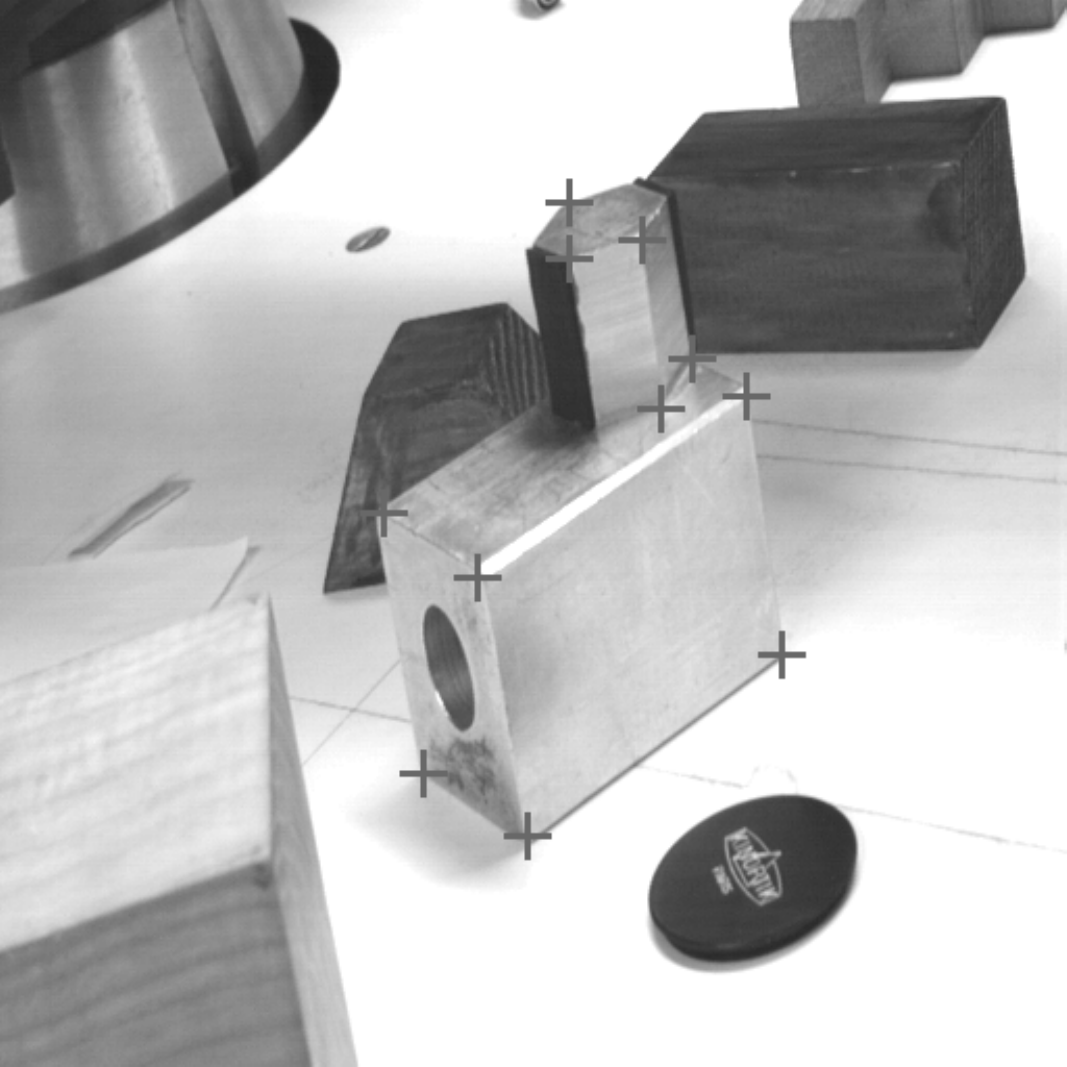}
\includegraphics[width=0.49\columnwidth]{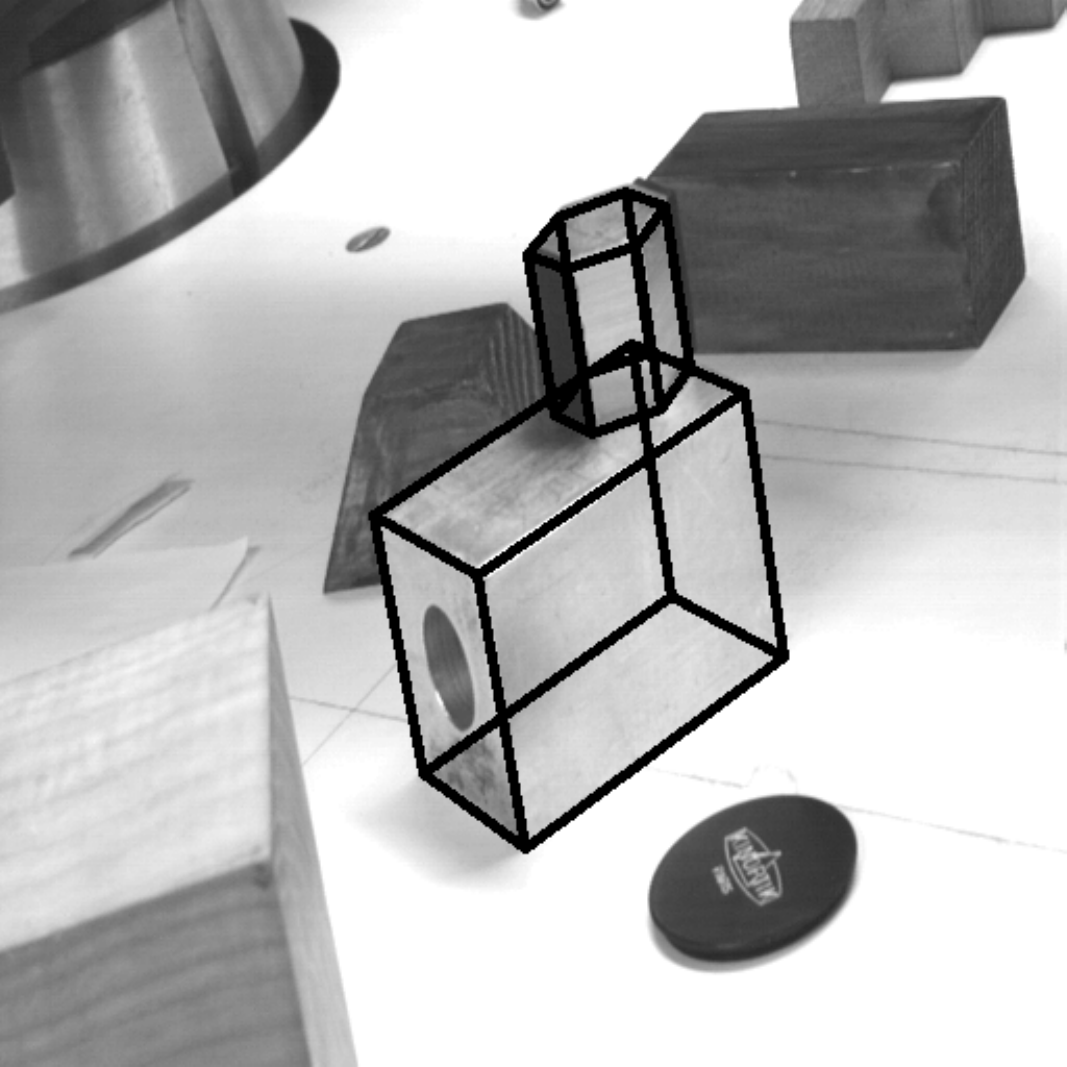}
\caption{It takes 1.47 miliseconds on a Sun-Sparc10/51 to compute the pose in
this case with the iterative paraperspective algorithm.}
\label{fig:11points}
\end{figure}

The fourth example illustrates the behaviour of the iterative paraperspective
and non linear algorithms as a function of the number of point
correspondences. As in the previous experiment we are interested by the speed
of convergence of the algorithm and by the computation time.
Figure~\ref{fig:11points}-left shows the image of a scene and 11 points that
correspond to the vertices of a polyhedral object.
Figure~\ref{fig:11points}-right shows the pose found for this object using
these 11 points and the iterative paraperspective algorithm.

Table~\ref{table:11points} summarizes the results obtained with a varying
number of point correspondences.

\begin{table*}
\begin{center}
\begin{tabular}{||c|c|c|c|c||}
\hline
	& \multicolumn{2}{c|}{Iterative paraperspective}& 
					\multicolumn{2}{c||}{Non linear} \\
\hline
Number of points & N. of iterations & CPU time (ms) & N. of iterations & CPU
time (ms) \\ \hline
4	& 5	& 0.87 	& 114 	& 87.5	\\ \hline
5	& 5	& 0.95	& 67	& 58.7	\\ \hline
6	& 6	& 1.02	& 66	& 68.0 	\\ \hline
7	& 5	& 1.06	& 66	& 72.5	\\ \hline
8	& 6	& 1.12	& 58	& 70.5	\\ \hline
9	& 6	& 1.33	& 82	& 108.6 \\ \hline
10	& 6	& 1.38 	& 78	& 117.4 \\ \hline
11	& 6	& 1.47	& 78	& 123.6 \\ \hline
\end{tabular}
\end{center}
\caption{A comparison of the iterative paraperspective and non linear methods
as a function of the number of points. One may notice the large number of
iterations required by the non linear method for 4 point correspondences.}
\label{table:11points}
\end{table*}

\section{Discussion}
\label{section:Discussion}

In this paper we proposed an extension to paraperspective of the iterative
weak perspective pose algorithm developped by DeMenthon \& Davis
\cite{DementhonDavis92}. We
established a link between perspective, weak perspective, and paraperspective 
camera models. We described a linear method for computing object
pose with a paraperspective model and an iterative algorithm which computes
pose with a perspective model by successive paraperspective approximations. 

We studied, both theoretically and experimentally, the convergence of the iterative
weak and paraperspective algorithms. We showed that, on an average, the latter
algorithm requires 2.5 times less iterations than the former algorithm.
Moreover, when the object is relatively close to the camera and at some 
distance 
from the optical axis, then the chance of
convergence of the paraperspective algorithm is higher than the chance of
convergence of the weak perspective one. Because it requires, on an
average, less iterations, the
iterative paraperspective algorithm is more time-efficient than the iterative weak
perspective algorithm. Of course, if the experimental conditions are such that the 
image center of gravity of the set of points is near the optical axis then the
weak perspective model should be prefered.

We showed that both algorithms (iterative weak and paraperspective) can take advantage
of a simple orthogonality constraint associated with the rotation
matrix of the object pose. 

Moreover, we compared the accuracy of the results obtained with these iterative linear
algorithms with the results obtained with a non-linear one. The former
algorithms are much faster and almost as precise as the latter, especially when the
orthogonality constraint mentioned above is being used. However, in the
presence of noise the performances of the iterative linear methods degrade faster with
increasing noise amplitude than the non linear method. This is due to the fact that
non-linear minimization numerical techniques are much more robust to noise than linear
algebraic techniques.

Whenever speed is an important concern, iterative linear pose methods should
however be prefered. Indeed, the iterative linear algorithms described by
DeMenthon and Davis \cite{DementhonDavis95} and in this paper are one hundred
times faster than the non-linear algorithms -- this feature allows one to include
these iterative linear techniques in real-time vision and robotics applications.

It is interesting to notice that the
iterative linear techniques described
here are able to deal with the problem of reconstruction from multiple views
\cite{ChristyHoraud94}. Indeed,
attempts have been made to solve the multiple view reconstruction problem
using either an approximated (linear) camera model or a projective
(non-linear) camera model, but no attempts have been made to properly establish
a link between these two classes of algorithms.

\balance
\bibliographystyle{plain}

\end{document}